\documentclass[letterpaper,journal]{IEEEtran}

\usepackage[utf8]{inputenc}
\usepackage[T1]{fontenc}
\usepackage{microtype}
\usepackage{graphicx}
\usepackage{amsmath,amssymb,amsfonts}
\usepackage{array}
\usepackage{booktabs}
\usepackage{multirow}
\usepackage[caption=false,font=normalsize,labelfont=sf,textfont=sf]{subfig}
\usepackage{stfloats}
\usepackage{placeins}
\usepackage[nocompress]{cite}
\usepackage{xcolor}
\usepackage{url}
\usepackage[
  hidelinks,
  pdftitle={Feed-Forward Multi-View Feature Coding for 3D Gaussian Splatting},
  pdfauthor={Seungmi Choi, Ji-Hyoun Choi, Suhyun Jo, Tae-Gwon Park, and Hui Yong Kim},
]{hyperref}
\usepackage[capitalize,noabbrev]{cleveref}
\crefname{figure}{Fig.}{Figs.}
\Crefname{figure}{Fig.}{Figs.}

\usepackage{soul}
\sethlcolor{yellow}
\soulregister\Cref7
\soulregister\cite7
\soulregister\constant0
\soulregister\adaptive0
\soulregister\method0
\newif\ifappendixhighlights
\appendixhighlightsfalse
\DeclareRobustCommand{\rev}[1]{\ifappendixhighlights\hl{#1}\else#1\fi}
\DeclareRobustCommand{\revheading}[1]{\ifappendixhighlights\colorbox{yellow}{#1}\else#1\fi}
\newif\ifreviewhighlights
\reviewhighlightsfalse
\DeclareRobustCommand{\reviewrev}[1]{\ifreviewhighlights\hl{#1}\else#1\fi}
\newif\ifrevfixhighlights
\revfixhighlightsfalse
\DeclareRobustCommand{\revfix}[1]{\ifrevfixhighlights\hl{#1}\else#1\fi}
\DeclareRobustCommand{\revfixmath}[1]{\ifrevfixhighlights{\setlength{\fboxsep}{0.5pt}\colorbox{yellow}{$#1$}}\else\ensuremath{#1}\fi}
\newcolumntype{L}[1]{>{\raggedright\arraybackslash}p{#1}}

\title{Feed-Forward Multi-View Feature Coding for 3D Gaussian Splatting}

\author{Seungmi Choi, Ji-Hyoun Choi, Suhyun Jo, Tae-Gwon Park, and Hui Yong Kim,~\IEEEmembership{Member,~IEEE}%
\thanks{This work was supported primarily by the National Research Foundation of Korea (NRF) grant funded by the Korean government (MSIT) in 2026 (Grant No. RS-2026-25617533, Project Title: Data-Efficient AI Architecture and Training/Inference Methods for Dynamic Modeling of 3D World). This research was partially supported by the ANCHOR program through the Gyeonggi ANCHOR Center, funded by the Ministry of Education(MOE) and the GYEONGGI-DO, Republic of Korea. (2026-ANCHOR-09-A07) This research is partially funded by the BK21 FOUR program of Graduate School, Kyung Hee University. (GSX-20260795)
\textit{(Seungmi Choi and Ji-Hyoun Choi are co-first authors.)
(Corresponding author: Hui Yong Kim.)}}%
\thanks{The authors are with the School of Computing, Kyung Hee University,
Yongin 17104, Republic of Korea (e-mail: 2018102242@khu.ac.kr;
jihyoun0403@khu.ac.kr; suhyun.jo@khu.ac.kr; clue9986@khu.ac.kr;
hykim.v@khu.ac.kr).}}

\newcommand{\method}{MFGS}
\newcommand{\constant}{MFGS-Base}
\newcommand{\adaptive}{MFGS-Adaptive}
\newcommand{\feat}{F}
\newcommand{\qfeat}{\hat{F}}
\newcommand{\gauss}{G}
\newcommand{\bits}{B}
\newcommand{\loss}{\mathcal{L}}

\newcommand{\lsim}{\mathcal{L}_{\mathrm{imp}}}
\newcommand{\inputset}{I_{\mathrm{in}}}
\newcommand{\targetset}{I_{\mathrm{tgt}}}

\begin{document}
\bstctlcite{IEEEexample:BSTcontrol}

\makeatletter
\def\@IEEEpubidpullup{2\baselineskip}
\makeatother
\IEEEpubid{\parbox[b]{\textwidth}{\centering\footnotesize This work has been submitted to the IEEE for possible publication.\\ Copyright may be transferred without notice, after which this version may no longer be accessible.}}

\maketitle

\begin{abstract}
Feed-forward 3D Gaussian Splatting (3DGS) methods enable efficient reconstruction of 3D scenes from sparse multi-view images in a single forward pass, without scene-specific optimization. However, existing feed-forward scene-coding pipelines follow a ``generate-then-compress'' framework, first generating a Gaussian representation and then encoding its geometry and appearance through separate coding paths, complicating codec design and limiting the exploitation of cross-attribute redundancy. We propose multi-view feature coding for Gaussian splatting (\method{}), an end-to-end feed-forward framework for scene coding. Instead of compressing an explicit Gaussian representation, \method{} extracts multi-view features from the input images and encodes them into a compact bitstream. Through end-to-end rate--distortion training, \method{} learns \revfix{a compact latent representation of multi-view features that jointly represents} geometry and appearance, enabling the receiver to predict Gaussian attributes from the decoded features. We further introduce importance-guided Gaussian selection to accommodate different rendering budgets, allowing the receiver to adjust the number of Gaussians used for rendering from the same bitstream. Across RE10K, ACID, and DL3DV, \method{} achieves $99.76$--$99.86\%$ bitrate savings over the generate-then-compress pipeline, as measured by BD-rate, while reducing total GMACs by $52$--$55\%$ and providing an approximately $12\times$ end-to-end speedup.
\end{abstract}

\begin{IEEEkeywords}
3D Gaussian Splatting (3DGS), 3D scene coding, feed-forward 3D reconstruction,
learned compression, multi-view feature coding.
\end{IEEEkeywords}

\section{Introduction}
\label{sec:intro}

\IEEEPARstart{3}{D} Gaussian Splatting (3DGS)~\cite{kerbl20233d} enables high-quality novel-view synthesis with real-time rendering, but its large explicit representations incur substantial storage and transmission costs. Optimization-based 3DGS compression methods~\cite{hac2024,hacpp2025,contextgs2024,compgs2024,rdogaussian2024}, which optimize Gaussian representations for each scene, achieve strong rate--distortion performance at the cost of time-consuming iterative optimization. This limits their applicability to live applications that require rapid reconstruction and transmission of newly captured scenes.

\begin{figure*}[!tp]
  \centering
  \includegraphics[width=\linewidth]{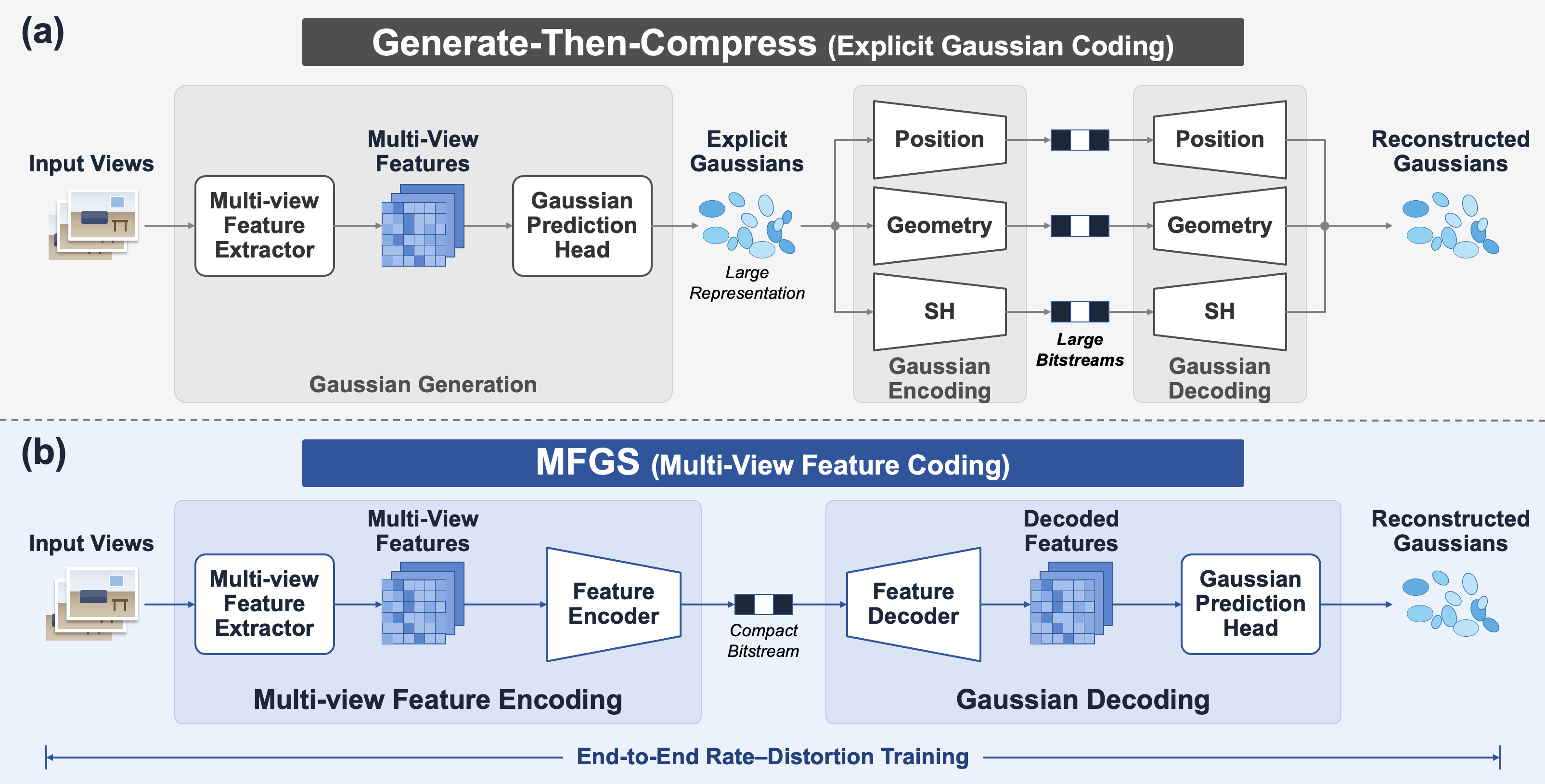}
  \caption{Comparison of 3D scene coding pipelines. (a) Generate-then-compress predicts explicit Gaussians and encodes their attributes through specialized paths, limiting joint compression of geometry and appearance. Non-differentiable operations such as G-PCC also hinder end-to-end training. (b) The \method{} framework compresses multi-view features with an established learned image codec and predicts Gaussian attributes from the decoded features at the receiver. This unified path simplifies coding, exploits cross-attribute redundancy, and enables end-to-end rate--distortion training, while compact latent grids reduce computation.}
  \label{fig:teaser}
\end{figure*}

Feed-forward methods~\cite{pixelsplat2023,mvsplat2024,depthsplat2024,monosplat2025,transplat2025} address this limitation by reconstructing Gaussian representations directly from sparse input views in a single pass. Scene transmission then follows the generate-then-compress pipeline in \Cref{fig:teaser}, which encodes the predicted Gaussian attributes through specialized coding paths. For example, FCGS~\cite{fcgs2025}, a representative feed-forward 3DGS compression model, uses G-PCC~\cite{schwarz2019emerging} for positions and learned coding for the remaining attributes. This separation complicates the architecture and limits joint compression of shared geometry and appearance information, while non-differentiable operations hinder end-to-end rate--distortion training. FCGS also requires repeated spatial and channel context evaluations over per-Gaussian latents and sequential decoding across \revfix{spatial and channel groups}, increasing coding latency.

\begin{figure}[!t]
  \centering
  \includegraphics[width=\linewidth]{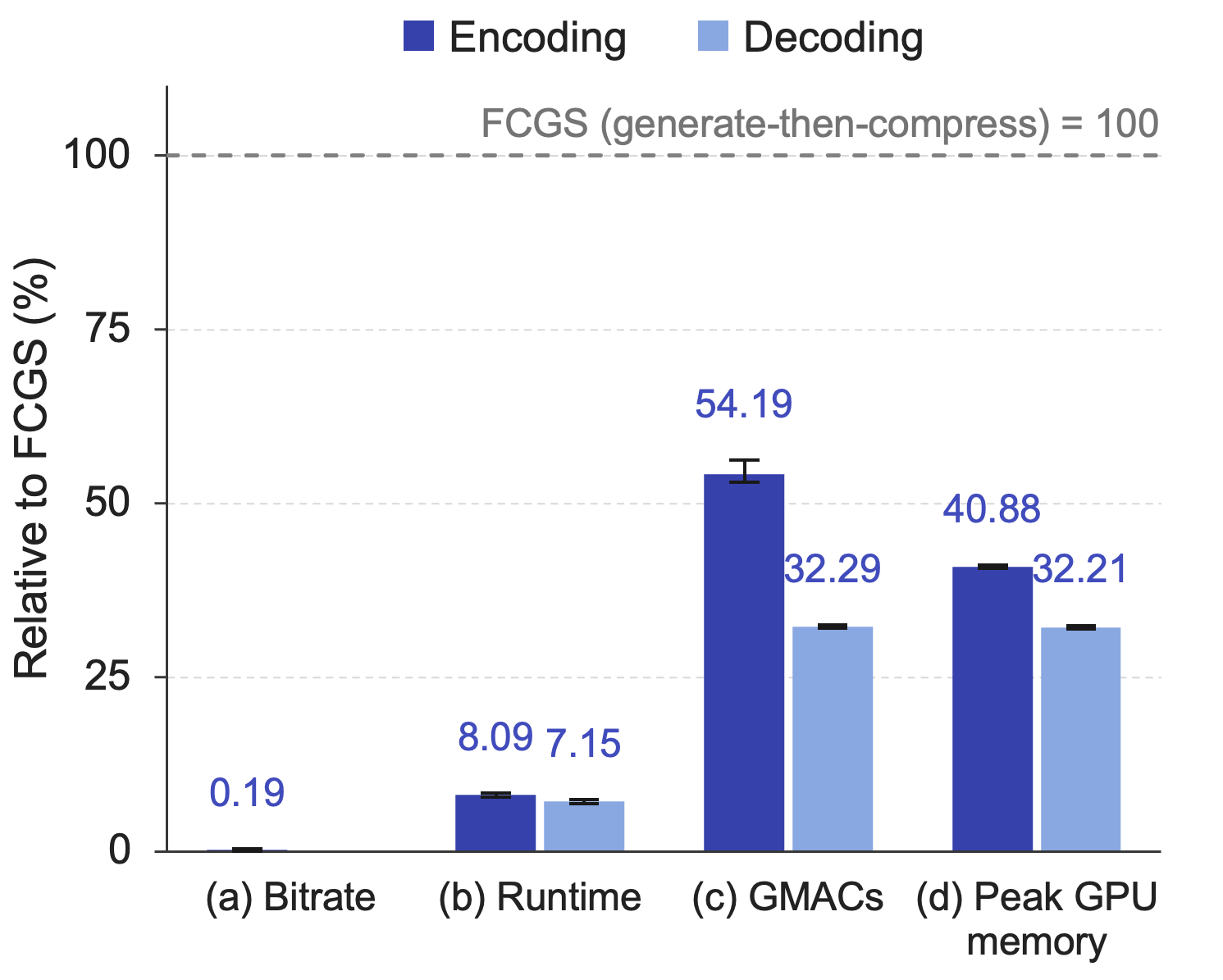}
  \caption{Relative bitrate and complexity of \textbf{\constant{}} versus FCGS (generate-then-compress), using the same generator architecture (\Cref{sec:base_results}). FCGS is normalized to $100\%$; lower is better. (a) Bitrate at comparable quality, (b) runtime, (c) GMACs, and (d) peak GPU memory. Panels (b)--(d) distinguish encoding and decoding. Bars show means across datasets; whiskers span their minima and maxima.}
  \label{fig:teaser_runtime}
\end{figure}

\IEEEpubidadjcol
We propose the multi-view feature coding for Gaussian splatting (\method{}) framework for end-to-end feed-forward 3D scene coding. Our base model, \textbf{\constant{}}, encodes multi-view features and predicts all Gaussian attributes from the decoded features at the receiver (\Cref{fig:teaser}). The regular spatial structure of these features allows the use of an established learned image codec, simplifying codec design. Their joint representation of geometry and appearance enables end-to-end rate--distortion optimization to exploit cross-attribute redundancy. This approach improves rate--distortion performance while reducing computation, memory usage, and runtime relative to the generate-then-compress baseline (\Cref{fig:teaser_runtime}), providing a basis for efficient 3D live streaming.

Receiver devices vary in computational capacity, while interactive viewing requires rendering within a per-frame time budget. To accommodate these constraints, our adaptive model, \textbf{\adaptive{}}, combines importance-guided bit allocation with receiver-side Gaussian selection. At each rate point, the receiver can adjust the number of rendered Gaussians from a single bitstream to match its rendering capacity.

Our main contributions are summarized as follows:
\begin{itemize}
  \item We introduce the \textbf{\method{} framework} as an alternative to generate-then-compress for feed-forward 3D scene coding. By \revfix{placing the coding interface before Gaussian prediction, the framework replaces attribute-specific coding paths with a single learned image-coding path}, enabling joint compression of geometry and appearance and end-to-end rate--distortion optimization.
  \item We instantiate the proposed framework in the \textbf{\constant{}} model and demonstrate substantial bitrate savings together with lower computation, peak GPU memory, and runtime than the generate-then-compress baseline across RE10K, ACID, and DL3DV. These gains substantially reduce both transmission and processing costs, enabling more efficient delivery and reconstruction of 3D scenes.
  \item We extend \revfix{MFGS-Base} with the \textbf{\adaptive{}} model, enabling receiver-side adjustment of the number of rendered Gaussians from a single bitstream. Experiments demonstrate limited quality degradation across rendering budgets, allowing a shared bitstream to serve receivers with different rendering capacities.
\end{itemize}

\section{Related Work}
\label{sec:related}

\subsection{Optimization-based 3DGS compression}
Optimization-based 3DGS compression methods optimize a compact Gaussian representation separately for each scene. HAC~\cite{hac2024} models the relationship between unstructured anchors and a structured hash grid, while HAC++~\cite{hacpp2025} further captures dependencies among attributes within each anchor. ContextGS~\cite{contextgs2024} introduces a hierarchical autoregressive context model across anchor levels. CompGS~\cite{compgs2024} applies K-means-based vector quantization while optimizing the Gaussian parameters, whereas RDO-Gaussian~\cite{rdogaussian2024} combines pruning and entropy-constrained vector quantization under a rate--distortion objective. These methods achieve strong compression performance but require scene-specific optimization, limiting their use in live applications.

\subsection{Feed-forward 3DGS generation}
Feed-forward 3DGS methods predict explicit Gaussian primitives from input multi-view images without per-scene optimization. pixelSplat~\cite{pixelsplat2023} models Gaussian depth using a probabilistic parameterization, while MVSplat~\cite{mvsplat2024} estimates depth using plane-sweep cost volumes. DepthSplat~\cite{depthsplat2024} further combines multi-view cost volumes with features from a pretrained monocular depth model. MonoSplat~\cite{monosplat2025} adapts monocular depth features into multi-view representations, whereas TranSplat~\cite{transplat2025} combines confidence-guided feature matching with monocular depth priors. ZPressor~\cite{zpressor2025} aggregates information from support views into a compact anchor-view latent representation, improving scalability to dense input views.

\subsection{Feed-forward 3DGS compression}
FCGS~\cite{fcgs2025} encodes an existing Gaussian representation without scene-specific fine-tuning using a multi-path entropy module and inter- and intra-Gaussian context models. The MPEG-GSC test model TMCV~\cite{mpeggsc2025} also provides coding tools for explicit Gaussian parameters. These codecs retain the generated Gaussians as the coded representation, so coding operates on explicit geometry and appearance attributes. In contrast, our approach moves coding before Gaussian prediction and jointly compresses the shared information in an image-like multi-view feature representation.

\subsection{Gaussian pruning}
Gaussian pruning reduces the number of primitives, thereby lowering storage and rasterization costs. LightGaussian~\cite{lightgaussian2023} ranks Gaussians using a global significance score computed over the training views and removes the least significant primitives. It then recovers rendering quality through SH distillation with pseudo views and applies vector quantization. PUP-3DGS~\cite{pup3dgs2024} uses a second-order approximation of reconstruction sensitivity with respect to the spatial parameters of each Gaussian and applies a multi-round prune--refine procedure. EcoSplat~\cite{ecosplat2025} introduces a two-stage training scheme for feed-forward 3DGS. It first learns pixel-aligned Gaussian prediction and then ranks and adjusts the predicted primitives according to a target primitive count.

\subsection{Learned image compression}
Learned image compression uses neural analysis and synthesis transforms, quantized latent representations, and entropy models trained under a rate--distortion objective. The scale hyperprior model~\cite{balle2018variational} predicts spatially varying scale parameters for the latent distribution. Joint autoregressive and hierarchical priors~\cite{minnen2018joint} combine a hierarchical prior with causal autoregressive context to improve entropy modeling. Attention-based image compression~\cite{cheng2020learned} uses discretized Gaussian mixture likelihoods to model the latent distribution. AuxT~\cite{li2025disentangled} introduces wavelet-based linear shortcuts that support feature decorrelation and energy modulation, accelerating the training of nonlinear transforms.

\section{Method}
\label{sec:method}

\subsection{System formulation}
\label{sec:problem}

Let $\inputset=\{(I_i,\Pi_i)\}_{i=1}^{C}$ denote a set of $C$ sparse input views, where $I_i$ and $\Pi_i$ are the RGB image and camera parameters of view $i$, respectively. Given $\inputset$, our goal is to transmit a compact bitstream $\bits$ from which the receiver reconstructs a Gaussian representation $\hat{\gauss}$ for novel-view synthesis.

Conventional feed-forward compression follows a generate-then-compress pipeline:
\begin{equation*}
  \inputset
  \xrightarrow{\substack{\text{multi-view}\\\text{feature extraction}}}
  \feat
  \xrightarrow{\substack{\text{Gaussian}\\\text{prediction}}}
  \gauss
  \xrightarrow{\substack{\text{Gaussian}\\\text{encoding}}}
  \bits
  \xrightarrow{\substack{\text{Gaussian}\\\text{decoding}}}
  \hat{\gauss} .
\end{equation*}
The sender generates an explicit Gaussian representation ($\inputset\!\to\!\gauss$) before encoding it into a bitstream ($\gauss\!\to\!\bits$). $\feat$ denotes the multi-view features extracted from $\inputset$ before Gaussian prediction.

The \method{} framework instead encodes the multi-view features and performs Gaussian prediction at the receiver:
\begin{equation*}
  \inputset
  \xrightarrow{\substack{\text{multi-view}\\\text{feature}\\\text{extraction}}}
  \feat
  \xrightarrow{\substack{\text{multi-view}\\\text{feature}\\\text{encoding}}}
  \bits
  \xrightarrow{\substack{\text{multi-view}\\\text{feature}\\\text{decoding}}}
  \qfeat
  \xrightarrow{\substack{\text{Gaussian}\\\text{prediction}}}
  \hat{\gauss} .
\end{equation*}
The receiver decodes the multi-view features from the bitstream ($\bits\!\to\!\qfeat$) before predicting the Gaussian attributes ($\qfeat\!\to\!\hat{\gauss}$). The transmitted representation can therefore be optimized jointly for the geometry and appearance needed for rendering.

\begin{figure*}[!tp]
  \centering
  \includegraphics[width=\linewidth]{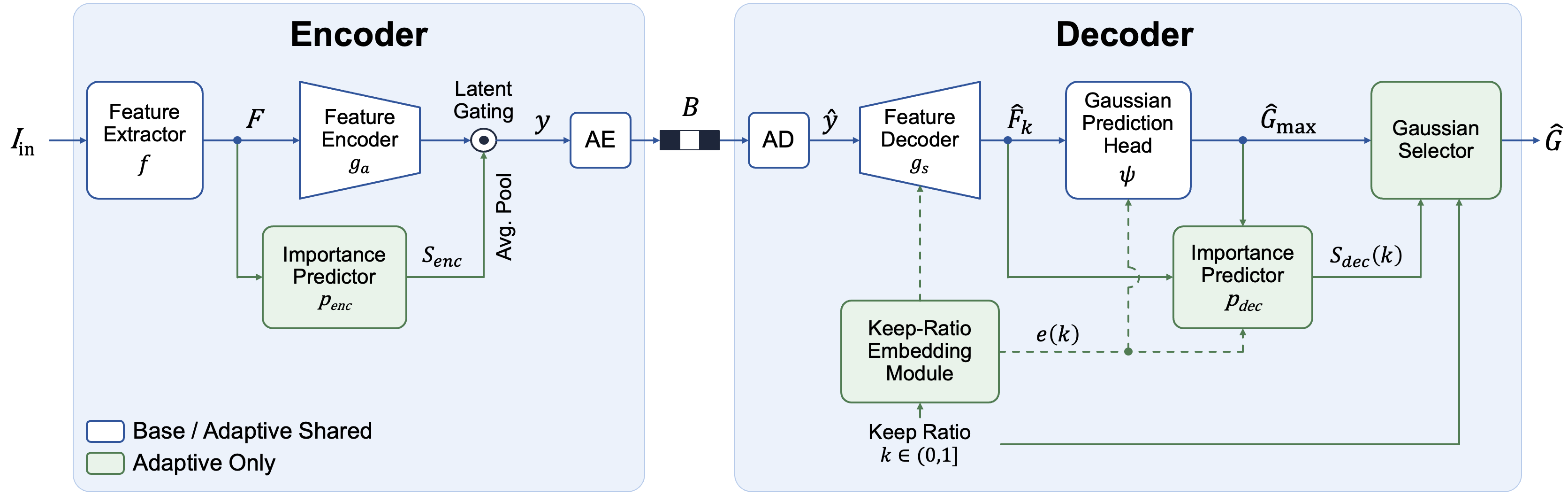}
  \caption{Overview of the \method{} framework from multi-view feature extraction through Gaussian reconstruction. The sender extracts $\feat$ from sparse input views and encodes it into a bitstream, while the receiver decodes $\qfeat$ and predicts Gaussian primitives. The \adaptive{} model additionally supports importance-guided bit allocation and receiver-side Gaussian selection. Green modules are specific to the \adaptive{} model. AE and AD denote arithmetic encoding and decoding, respectively.}
  \label{fig:method}
\end{figure*}

\Cref{fig:method} summarizes the two models within the \method{} framework. The \constant{} model implements end-to-end multi-view feature coding, while the \adaptive{} model extends it to support adjustable rendering budgets at the receiver.

\subsection{End-to-end multi-view feature coding (\constant{})}
\label{sec:architecture}
\label{sec:base}

The \constant{} model is designed to improve compression efficiency while simplifying the coding process. Whereas generate-then-compress pipelines encode the attributes of spatially unstructured Gaussians through separate coding paths, \constant{} encodes multi-view features on regular image-like grids. This representation supports a unified codec whose entropy model operates on compact, downsampled latents, reducing computation and memory usage. End-to-end rate--distortion training further enables the codec to exploit redundancy between geometry and appearance within this unified coding path.

\subsubsection{Gaussian generator and multi-view features}
We instantiate the generator with MVSplat~\cite{mvsplat2024}, extended with ZPressor~\cite{zpressor2025}. Regardless of the number $C$ of input views, ZPressor aggregates them into a fixed number $V$ of cluster-view planes, forming the multi-view features $\feat$, each plane of which has the spatial dimensions $H\times W$ of the input multi-view images. Since the Gaussian prediction head produces one Gaussian per spatial location, this yields $N=VHW$ Gaussians, independent of $C$. We denote the generator's multi-view feature extractor and Gaussian prediction head by $f$ and $\psi$, respectively. The \constant{} model inserts a multi-view feature codec at the interface between them: the sender applies $f$ and encodes $\feat$, while the receiver decodes $\qfeat$ and applies $\psi$ to obtain $\hat{\gauss}=\psi(\qfeat)$ for rendering. More architectural details are deferred to \Cref{app:architecture}.

\subsubsection{Multi-view feature codec and entropy model}
Each multi-view feature plane is a three-dimensional tensor indexed by channel, height, and width, with the same regular spatial layout as an image. This allows us to use an established learned image codec, specifically the mean-scale hyperprior (MSH) of~Minnen \emph{et al.}~\cite{minnen2018joint}, extended with AuxT~\cite{li2025disentangled}. To accommodate the multi-view features, we adjust the input channels of the analysis transform $g_a$ and the output channels of the synthesis transform $g_s$ from three RGB channels to 133 feature channels. We apply the same codec to all feature planes and train it under the end-to-end rate--distortion objective.

The analysis transform $g_a$ maps $\feat$ to a latent $y$, and the hyper-analysis transform $h_a$ maps $y$ to a hyper-latent $z$. Entropy coding operates on the quantized latents, where $Q$ denotes quantization:
\begin{equation}
  y=g_a(\feat), \quad
  z=h_a(y), \qquad
  \hat y=Q(y), \quad
  \hat z=Q(z).
\end{equation}
The hyper-synthesis transform $h_s$ predicts the parameters of the conditional model on $\hat y$,
\begin{equation}
  (\mu,\sigma)=h_s(\hat z), \qquad
  p_{\hat y\mid\hat z}(\hat y\mid\hat z)
  =\prod_j p_j(\hat y_j\mid\mu_j,\sigma_j),
\end{equation}
where each $p_j$ is the probability mass of a discretized Gaussian conditional.
At inference, $\hat z$ is coded under a learned entropy bottleneck $p_{\hat z}$, and $\hat y$ under $p_{\hat y\mid\hat z}$. The synthesis transform $g_s$ then reconstructs the multi-view features, and the prediction head produces the Gaussians used for rendering,
\begin{equation}
  \qfeat=g_s(\hat y), \qquad
  \hat{\gauss}=\psi(\qfeat).
\end{equation}
The bitstream includes both latent streams, cluster-view camera parameters, depth bounds, and tensor-shape metadata. MFGS reconstruction quality is evaluated after decoding the stored bitstream, and rate is measured from its complete file size. Layer-level codec details are deferred to \Cref{app:architecture}.

\subsubsection{Training objective}
\constant{} is trained end-to-end with a rate--distortion loss. The rate term is
\begin{equation}
  R=\frac{1}{\revfixmath{NA}}\,\mathbb{E}\!\left[
    -\log_2 p_{\hat y\mid\hat z}(\hat y\mid\hat z)
    -\log_2 p_{\hat z}(\hat z)
  \right].
\end{equation}
The log-probability terms account for all quantized symbols in the scene bitstream, and \revfix{the rate is normalized by the total number of Gaussian attributes, where $A$ is the number of scalar attributes per Gaussian.} During training, we use a set of $U$ target image--camera pairs, $\targetset=\{(I_u,\Pi_u)\}_{u=1}^{U}$, for rendering supervision, where $I_u$ is the ground-truth image observed from target camera $\Pi_u$. \revfix{For the reconstructed Gaussian representation $\hat{\gauss}$, the distortion is measured in image space as}
\begin{equation}
  \begin{aligned}
  D(\revfixmath{\hat{\gauss}})=\frac{1}{U}\sum_{u=1}^{U}\Bigl[
    &\operatorname{MSE}\!\left(\mathcal{R}(\revfixmath{\hat{\gauss}},\Pi_u),I_u\right) \\
    &+\alpha\,\operatorname{LPIPS}\!\left(\mathcal{R}(\revfixmath{\hat{\gauss}},\Pi_u),I_u\right)
  \Bigr],
  \end{aligned}
\end{equation}
where $\mathcal{R}$ denotes differentiable Gaussian rasterization and $\alpha$ weights the LPIPS perceptual term~\cite{zhang2018lpips}. \constant{} renders the full decoded set, so its objective reduces to
\begin{equation}
  \loss_{\mathrm{base}}=D(\hat{\gauss})+\lambda R,
\end{equation}
where $\lambda$ controls the operating point. The rate term and rendering loss jointly supervise the information used to predict geometry and appearance, allowing the multi-view feature representation and codec to adapt together.

\begin{figure*}[!t]
  \centering
  \includegraphics[width=0.82\linewidth]{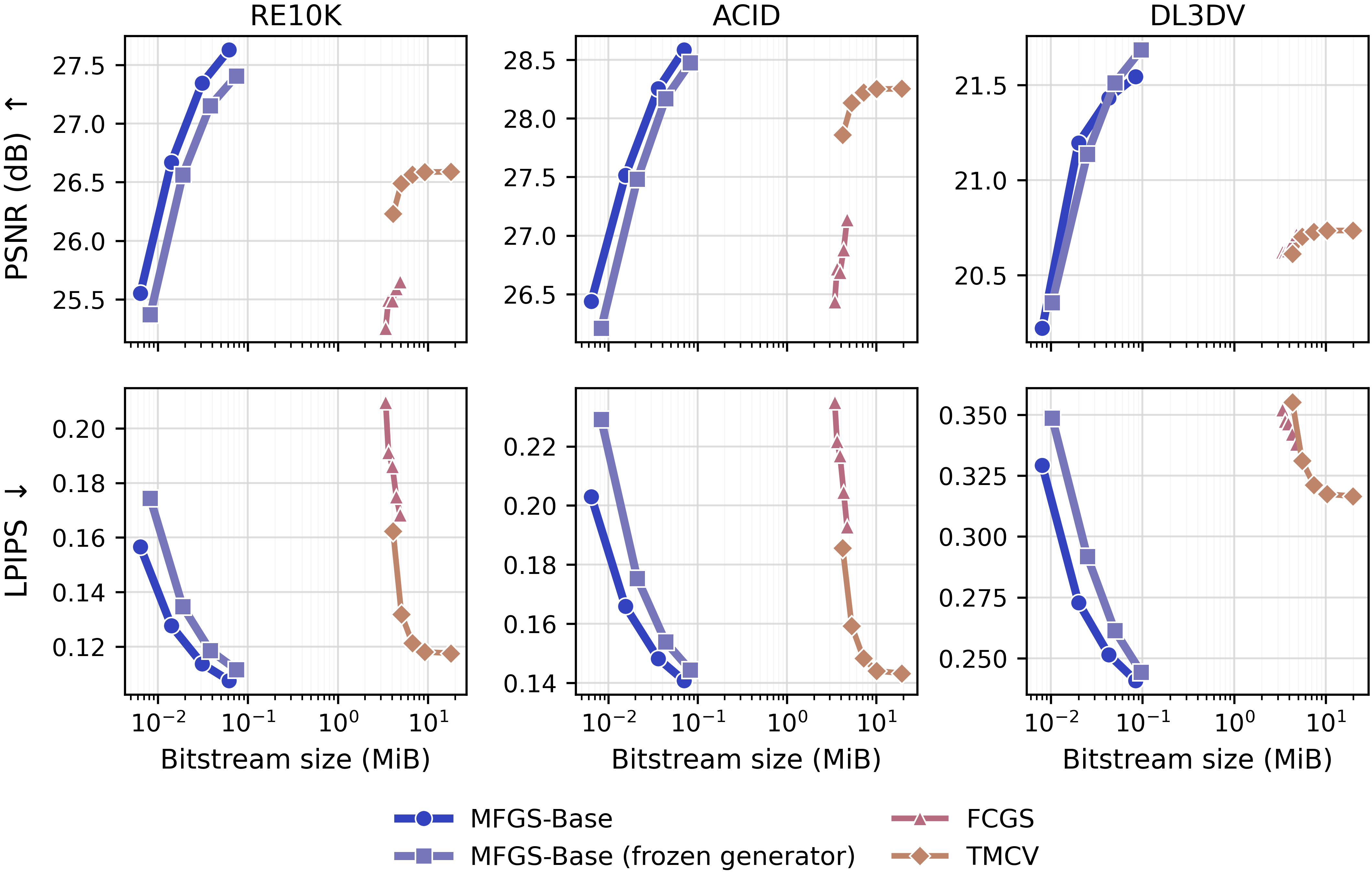}
  \caption{Rate--distortion curves for the \constant{} model, its frozen-generator variant, FCGS, and TMCV on RE10K, ACID, and DL3DV, \revfix{with distortion measured by PSNR (top) and LPIPS (bottom)}. The horizontal axis shows scene bitstream size on a logarithmic scale.}
  \label{fig:rd}
\end{figure*}

\subsection{Decode-time budget adaptation (\adaptive{})}
\label{sec:adaptive}

The \adaptive{} model extends \constant{} to accommodate receivers with different rendering capacities. Unlike post-hoc pruning, it jointly optimizes feature coding, budget-conditioned reconstruction, and Gaussian selection using the additional modules shown in green in \Cref{fig:method}. Training across keep ratios $k\in(0,1]$, \revfix{sampled from $[k_{\min},1]$ with $k_{\min}=0.1$,} limits quality degradation as the rendering budget decreases. At each trained rate point, the bitstream remains independent of $k$, allowing the receiver to \reviewrev{adjust the Gaussian count after transmission}.

\subsubsection{Keep-ratio-conditioned reconstruction}
The keep ratio $k$ is mapped to an embedding $e(k)$ using a sinusoidal encoding~\cite{vaswani2017attention} followed by a two-layer MLP. Through feature-wise linear modulation (FiLM)~\cite{perez2018film}, $e(k)$ conditions the synthesis transform and Gaussian prediction head:
\begin{equation}
    \qfeat_k=g_s(\hat y,e(k)), \qquad
    \hat{\gauss}_{\max}=\psi(\qfeat_k,e(k)).
\end{equation}
The same embedding also conditions the decoder-side importance predictor described below.
The prediction head produces the full, maximum-budget candidate set $\hat{\gauss}_{\max}$, containing $N=VHW$ Gaussians. The receiver retains $K=\lfloor kN\rfloor$ of these candidates for the requested rendering budget.

\subsubsection{Importance-guided coding and selection}
Separate importance predictors guide bit allocation at the sender and candidate ranking at the receiver:
\begin{equation}
  S_{\mathrm{enc}}=p_{\mathrm{enc}}(\feat), \qquad
  S_{\mathrm{dec}}(k)
  =
  p_{\mathrm{dec}}
  \left(
    \qfeat_k,
    \hat{\gauss}_{\max},
    e(k)
  \right),
\end{equation}
where $S_{\mathrm{enc}},S_{\mathrm{dec}}(k)\in[0,1]^{V\times H\times W}$. Since $y$ has a lower spatial resolution than $S_{\mathrm{enc}}$, we apply the pooled encoder map as a latent gate, $y\leftarrow y\odot\operatorname{pool}(S_{\mathrm{enc}})$, before quantization and entropy coding. This gating encourages importance-dependent bit allocation under the rate--distortion objective. The importance map itself is not transmitted.

After decoding, the receiver predicts $S_{\mathrm{dec}}(k)$ and retains the $K$ highest-ranked candidates:
\begin{equation}
  \revfixmath{\hat{\gauss}(k)}
  =
  \operatorname{Top}_{K}
  \left(
    \hat{\gauss}_{\max},
    S_{\mathrm{dec}}(k)
  \right),
  \qquad
  K=\lfloor kN\rfloor.
\end{equation}
\revfix{Training uses this hard top-$K$ selection in the forward pass with a sigmoid straight-through gradient, and inference applies the same selection.}

\subsubsection{Training objective}
The encoder-side map is learned through the gated rate--distortion path. To train the receiver-side predictor, we treat this map as a stop-gradient target and sample $k\sim p(k)$ during training:
\begin{equation}
\begin{aligned}
  \lsim(k) &= \frac{1}{N}\left\|
    S_{\mathrm{dec}}(k)-\operatorname{sg}(S_{\mathrm{enc}})
  \right\|_2^2,\\
  \loss_{\mathrm{adapt}} &=
  \mathbb{E}_{k\sim p(k)}\!\left[
    D(\revfixmath{\hat{\gauss}(k)})+\beta\lsim(k)
  \right]+\lambda R,
\end{aligned}
\end{equation}
where $\operatorname{sg}(\cdot)$ denotes the stop-gradient operator and $\beta$ weights importance alignment. The objective evaluates distortion after top-$K$ selection and trains the model across a range of rendering budgets. Training and architecture details are provided in \revfix{\Cref{app:architecture}}.

\begin{table}[!t]
  \caption{BD-rate (\%) relative to FCGS (lower is better), using identical scene sets across all methods and rate points within each dataset (\Cref{app:evaluation_protocols}). The \adaptive{} model is evaluated at the full Gaussian budget ($k=1$).}
  \label{tab:bd_complexity}
  \centering
  \begin{tabular}{@{}lrrr@{}}
    \toprule
    Method & RE10K & ACID & DL3DV \\
    \midrule
    FCGS & $0.00$ & $0.00$ & $0.00$ \\
    TMCV & -- & -- & $+14.59$ \\
    \midrule
    \constant{} & $-99.86$ & $-99.80$ & $-99.76$ \\
    \adaptive{} ($k=1.0$) & $-99.84$ & $-99.80$ & $-99.77$ \\
    \bottomrule
  \end{tabular}
  \par\vspace{3pt}
  {\footnotesize --: no common PSNR interval for TMCV on RE10K/ACID.}
\end{table}

\section{Experiments}
\label{sec:experiments}

\subsection{Experimental setup}

We train the \constant{} and \adaptive{} models on RealEstate10K (RE10K)~\cite{re10k2018} using 12 input views and 8 target views per scene at a resolution of $256\times256$. 
We evaluate the trained checkpoints in-domain on RE10K and conduct cross-dataset evaluation without dataset-specific fine-tuning on ACID~\cite{acid2020} and DL3DV~\cite{dl3dv2024}, using 8 input views for RE10K and ACID and 12 for DL3DV, with 8 target views for all three datasets.
We train each model separately for each $\lambda\in\{0.0256,0.1024,0.4096,1.6384\}$. By default, we use 6 cluster views, \revfix{so that MFGS-Base and the generate-then-compress baselines reconstruct} 393\,216 Gaussians per scene. Additional evaluation details are provided in \Cref{app:evaluation_protocols}.

\subsection{\constant{} model results}
\label{sec:base_results}

\subsubsection{Comparison with generate-then-compress baselines}

We compare the \constant{} model with two generate-then-compress baselines, FCGS and the MPEG-GSC test model TMCV. Both compress the explicit output of a standalone MVSplat+ZPressor Gaussian generator, trained without a codec.

\constant{} achieves reconstruction quality comparable to FCGS and TMCV at substantially lower bitrates (\Cref{fig:rd}). Relative to FCGS, its BD-rates~\revfix{\cite{bjontegaard2001calculation}} are $-99.86\%$, $-99.80\%$, and $-99.76\%$ on RE10K, ACID, and DL3DV, respectively (\Cref{tab:bd_complexity}). To assess the contribution of end-to-end optimization, we also train only the codec while keeping the Gaussian generator frozen. This variant retains most of the coding gain, with joint training providing further bitrate reductions of $33.9\%$, $29.5\%$, and $20.5\%$ on the three datasets, respectively. A comparison with multi-view image coding (MIV) is provided in \Cref{app:miv}. These results support multi-view features as an effective coding target and demonstrate the additional benefit of joint representation and codec optimization.

\subsubsection{Qualitative comparison}
\label{sec:base_qualitative}

\begin{figure*}[!tp]
  \centering
  \includegraphics[width=0.9\linewidth]{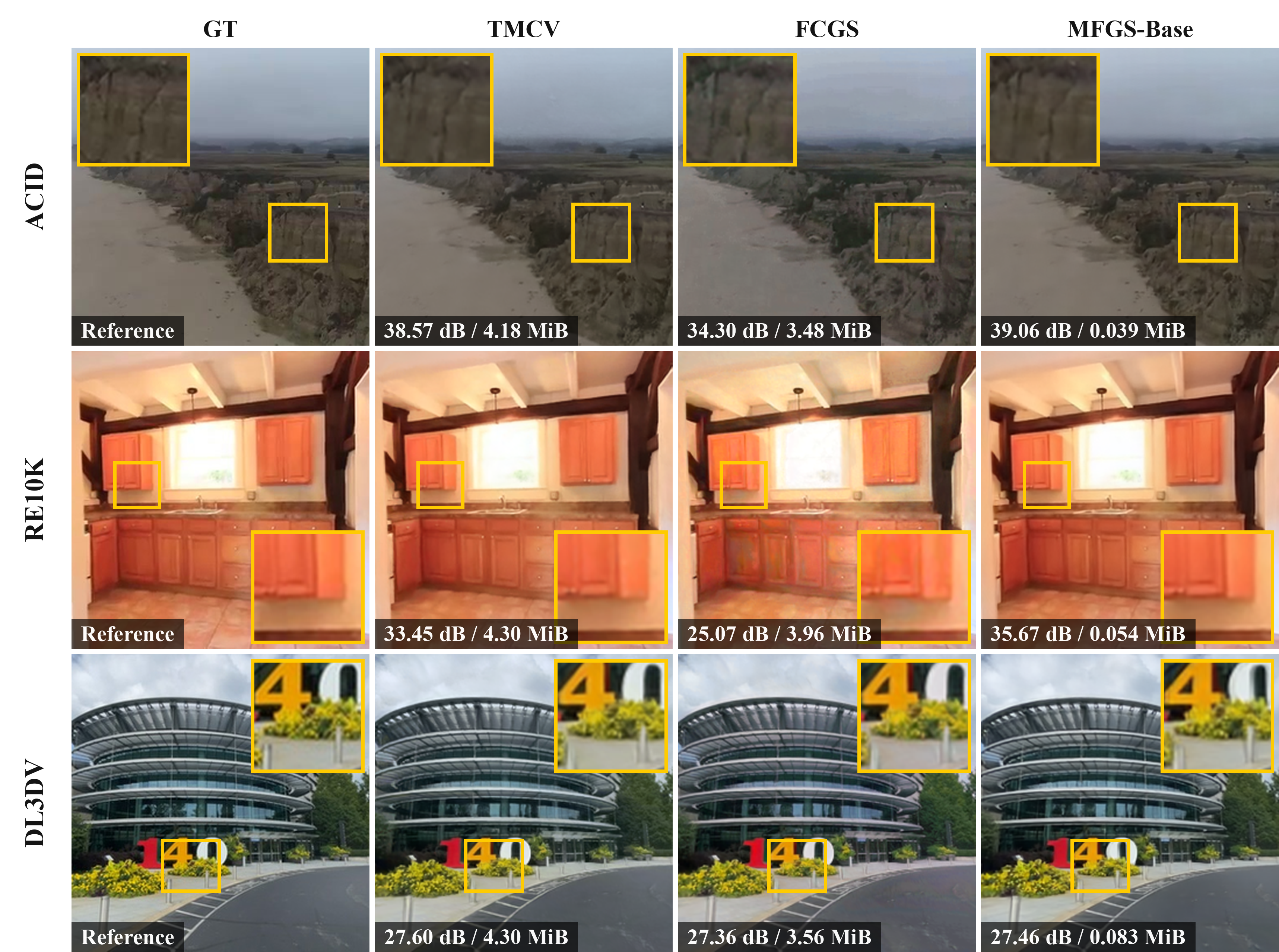}
  \caption{Target-view reconstructions on ACID, RE10K, and DL3DV. FCGS and TMCV compress the output of the standalone MVSplat+ZPressor generator. Yellow boxes mark regions enlarged in the insets. Each panel reports target-view PSNR and the complete scene bitstream size.}
  \label{fig:base_subjective_quality}
\end{figure*}

\reviewrev{\Cref{fig:base_subjective_quality} compares reconstructions at selected operating points.} We use the highest-rate trained setting for \constant{} ($\lambda=0.0256$) and the lowest-rate evaluated settings for FCGS ($\lambda=0.0016$) and TMCV (attribute QP~32, position QP~0). \reviewrev{At these settings, \constant{} \revfix{achieves visual quality comparable to both baselines while using} $43$--$107\times$ smaller bitstreams. It achieves higher target-view PSNR than both baselines in the ACID and RE10K examples, while the three methods have similar PSNR in the DL3DV example.}

\subsubsection{Complexity comparison}
We compare the complexity of the \constant{} model and MVSplat+FCGS (\Cref{tab:decoder_complexity}), with Gaussian generation by MVSplat+ZPressor included in the baseline encoding stage. For MVSplat+FCGS, runtime includes G-PCC execution, whereas all other reported complexity metrics---GMACs, peak GPU memory, and parameter counts---exclude G-PCC. \reviewrev{\revfix{Even with G-PCC excluded from the baseline,} \constant{} requires less than half the combined encoding and decoding GMACs and reduces peak GPU memory by approximately $60$--$70\%$ across the two stages, averaged over the three datasets.} \constant{} also achieves an approximately $12\times$ end-to-end speedup. These gains reflect the structure of entropy modeling in \constant{}, which predicts entropy parameters for all \revfix{latent} locations in a single pass over compact, downsampled multi-view grids. In contrast, FCGS requires repeated spatial and channel context evaluations over per-Gaussian latents and sequential decoding across spatial and channel groups. A module-wise analysis and the measurement protocol are provided in \Cref{app:runtime_protocol}.

\begin{table*}[!tp]
  \caption{Encoding and decoding complexity at $256\!\times\!256$ on RTX~3090 GPUs. Values are averaged across RE10K, ACID, and DL3DV. \revfix{Measurements use identical views from 100 randomly sampled test scenes from each of RE10K and ACID and all 140 DL3DV test scenes.} MVSplat+FCGS includes MVSplat+ZPressor generation. For MVSplat+FCGS, only runtime includes G-PCC execution; all other complexity metrics (GMACs, peak GPU memory, and parameters) exclude G-PCC. Parameters count learned weights only, in millions (M); G-PCC has no learned weights. Ratio denotes \textbf{\constant{}} divided by MVSplat+FCGS. Encoding and decoding exclude rendering. Operating points are $\lambda=0.0256$ for \textbf{\constant{}} and $\lambda=0.0016$ for MVSplat+FCGS.}
  \label{tab:decoder_complexity}
  \centering
  \small
  \setlength{\tabcolsep}{2pt}
  \begin{tabular*}{\linewidth}{@{}l@{\extracolsep{\fill}}rrrrrrrr@{}}
    \toprule
    & \multicolumn{4}{c}{Encoding} & \multicolumn{4}{c}{Decoding} \\
    \cmidrule(lr){2-5}\cmidrule(l){6-9}
    Method & GMACs$\downarrow$ & \shortstack{Peak\\(MiB)$\downarrow$} & \shortstack{Params.\\(M)} & \shortstack{Runtime\\(s)$\downarrow$} & GMACs$\downarrow$ & \shortstack{Peak\\(MiB)$\downarrow$} & \shortstack{Params.\\(M)} & \shortstack{Runtime\\(s)$\downarrow$} \\
    \midrule
    MVSplat+FCGS & 1601.30 & 8502.66 & 15.83 & 3.694 & 895.62 & 3971.57 & 2.19 & 2.683 \\
    \textbf{\constant{}} & \textbf{868.89} & \textbf{3475.75} & \textbf{27.78} & \textbf{0.299} & \textbf{289.17} & \textbf{1279.27} & \textbf{13.75} & \textbf{0.192} \\
    \midrule
    Ratio & $0.543\times$ & $0.409\times$ & $1.75\times$ & $0.081\times$ & $0.323\times$ & $0.322\times$ & $6.29\times$ & $0.072\times$ \\
    \bottomrule
  \end{tabular*}
  \vspace{8pt} 
\end{table*}

\subsubsection{Generalization across different Gaussian generators}
\begin{figure}[tp]
  \centering
  \includegraphics[width=\linewidth]{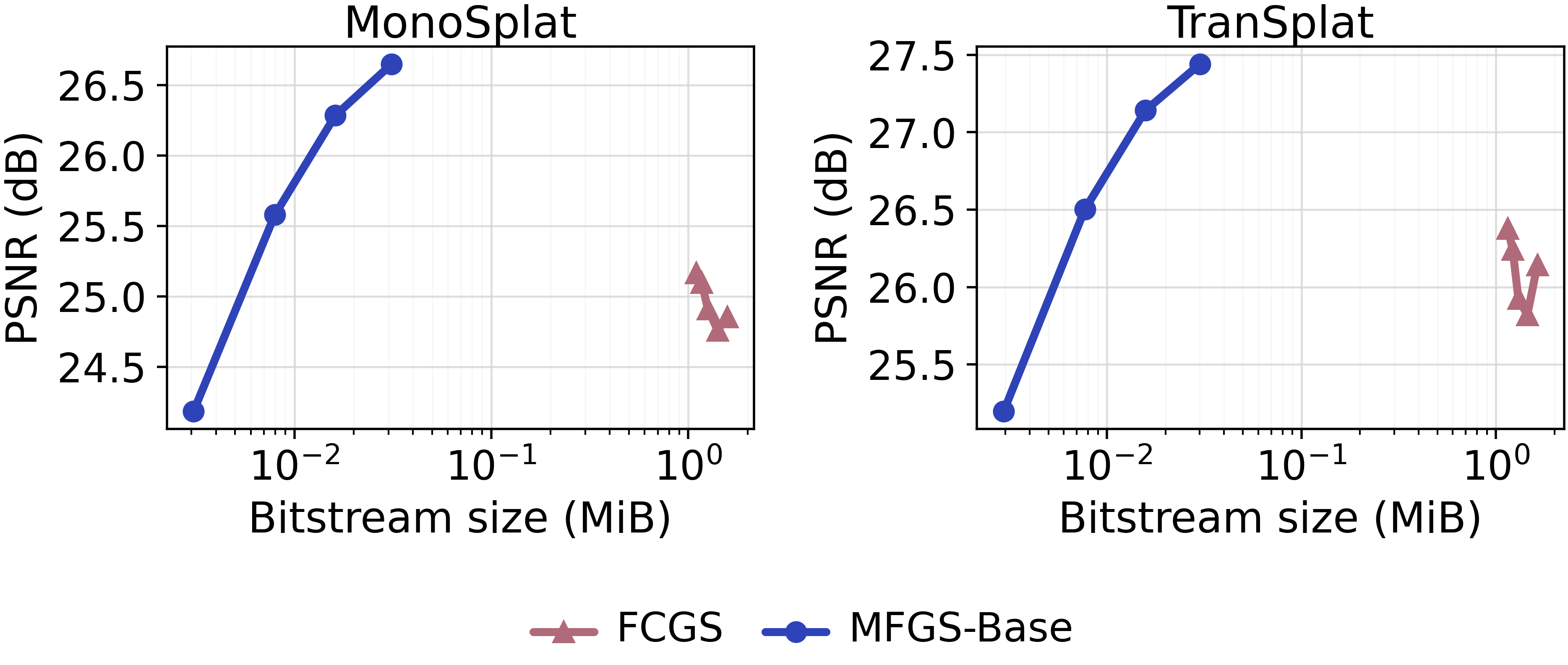}
  \caption{Rate--distortion comparison of the \constant{} model and matched FCGS baselines on ACID with MonoSplat and TranSplat.}
  \label{fig:backbone_generalization}
\end{figure}

\revfix{To assess the generalizability of the proposed framework across Gaussian generators, we conduct experiments with MonoSplat and TranSplat, which construct multi-view features differently from MVSplat. For each generator, we train an MFGS-Base model on RE10K and compare it on ACID with a generate-then-compress baseline that applies FCGS to the same generator's Gaussians, following the two-view setting used to train both generators (\Cref{app:evaluation_protocols}).} \constant{} achieves BD-rate reductions of $99.62\%$ and $99.60\%$ relative to these baselines with MonoSplat and TranSplat, respectively (\Cref{fig:backbone_generalization}). These consistent gains indicate that the coding advantage of the \method{} framework is not tied to a specific Gaussian generator.

\subsection{\adaptive{} model results}
\label{sec:adaptive_results}

\subsubsection{Comparison with post-hoc pruning}

\begin{figure}[tp]
  \centering
  \includegraphics[width=\linewidth]{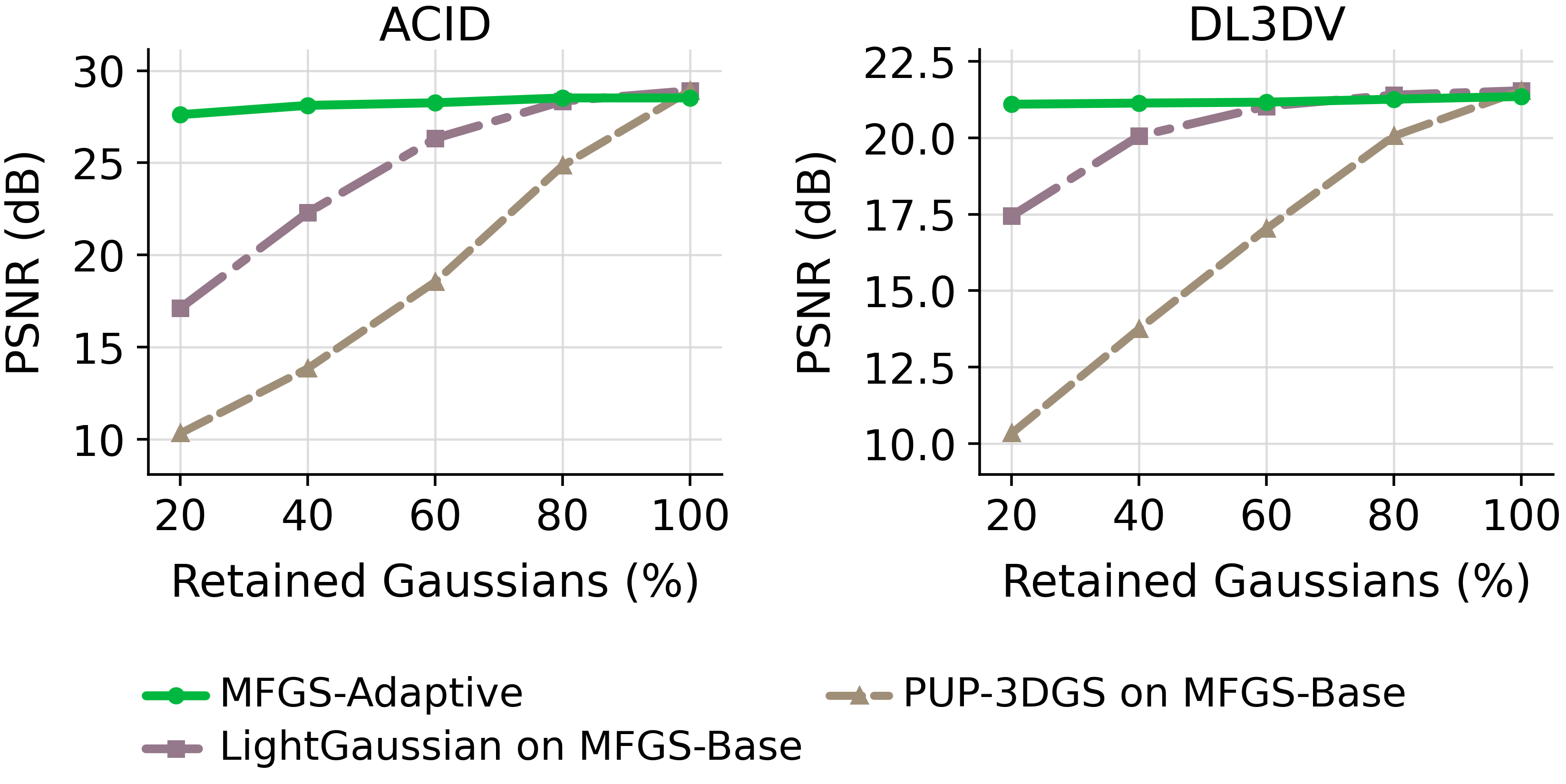}
  \caption{PSNR comparison between the \adaptive{} model and post-hoc pruning baselines across Gaussian keep ratios on ACID and DL3DV at $\lambda{=}0.0256$. LightGaussian and PUP-3DGS are applied to the Gaussians reconstructed by the \constant{} model.}
  \label{fig:pruning_aware_posthoc_rd}
\end{figure}

We compare the \adaptive{} model with the post-hoc pruning methods LightGaussian and PUP-3DGS. For a controlled comparison, both post-hoc methods rank the same Gaussians reconstructed by \constant{}, without retraining. As shown in \Cref{fig:pruning_aware_posthoc_rd}, \adaptive{} maintains substantially higher reconstruction quality as $k$ decreases. \revfix{Removing encoder-side latent gating and importance alignment together widens the PSNR range across keep ratios (\Cref{fig:gating_ablation}, \Cref{app:pruning_settings_ablation}).} These results indicate that the budget-conditioned architecture and training make reconstruction robust to changes in Gaussian count, limiting quality degradation across the evaluated rendering budgets.

\subsubsection{Qualitative comparison}
\begin{figure*}[!tp]
  \centering
  \includegraphics[width=0.9\linewidth]{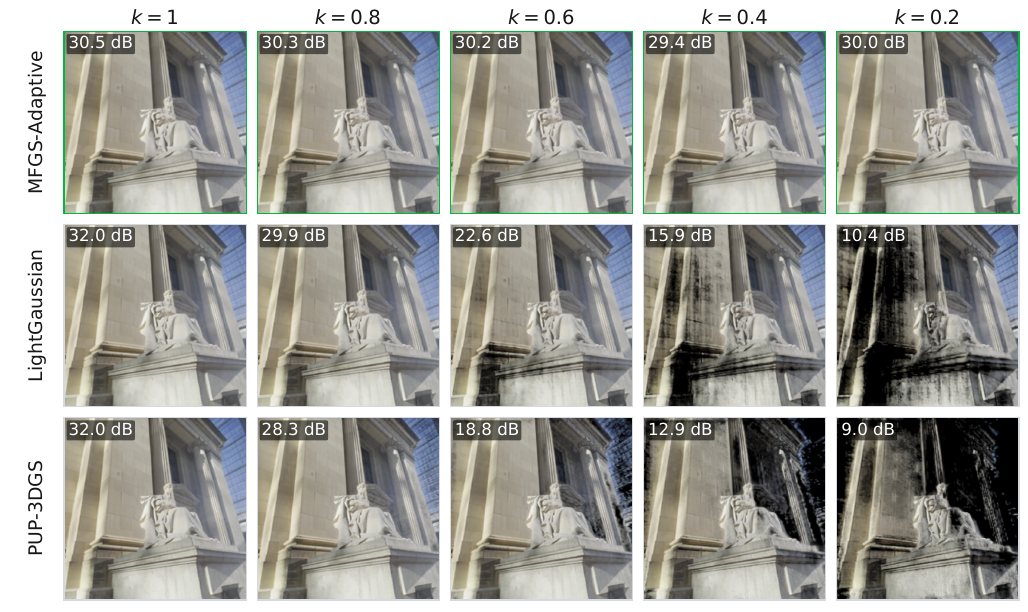}
  \caption{Qualitative pruning comparison on DL3DV across Gaussian keep ratios. Rows show the \adaptive{} model and post-hoc pruning baselines at $\lambda=0.0256$, and columns show $k$ decreasing from $1.0$ to $0.2$. Each panel reports PSNR with respect to the ground-truth target image. \reviewrev{The post-hoc baselines share the full-budget \constant{} reconstruction; \adaptive{} uses its own budget-conditioned reconstruction.}}
  \label{fig:pruning_subjective_quality_dl3dv}
\end{figure*}

\Cref{fig:pruning_subjective_quality_dl3dv} shows reconstructed views at different Gaussian keep ratios. The post-hoc baselines develop dark gaps and fragmented surfaces around the statue and facade at low keep ratios. \adaptive{} retains the main structure even at $k=0.2$.

\subsubsection{Scene-level Gaussian selection}
\label{sec:pruning_analysis}

\begin{figure*}[!tp]
  \centering
  \includegraphics[width=\linewidth]{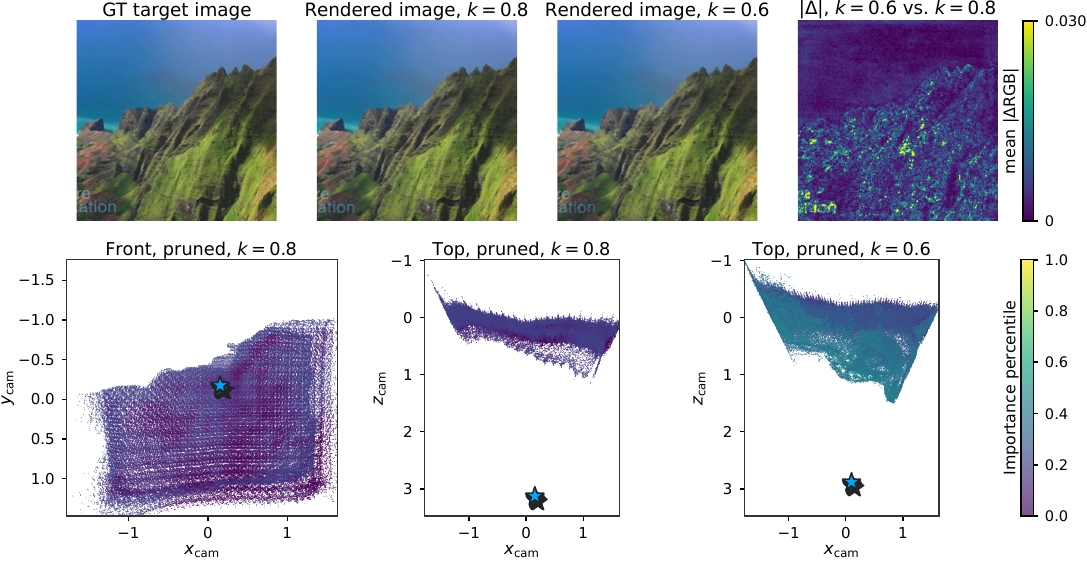}
  \caption{Scene-level Gaussian selection on ACID. The top row shows the ground-truth target image, reconstructions at $k{=}0.8$ and $k{=}0.6$ from the same bitstream, and their mean absolute RGB difference. The bottom row shows camera-aligned orthographic projections of excluded Gaussians, colored by decoder-side importance percentile within the full candidate set at each budget. Blue stars mark the reference input-camera center.}
  \label{fig:importance_geometry}
\end{figure*}

\reviewrev{In the ACID example in \Cref{fig:importance_geometry}, \adaptive{} preserves the overall scene structure across the two budgets,} \revfix{$k=0.8$ and $k=0.6$,} while excluded Gaussians cluster spatially and rendering changes concentrate along mountain ridges and slopes, with smaller differences over the sea. \revfix{Because $k$ also conditions reconstruction, the Gaussians retained at a lower budget are not constrained to be a subset of those retained at a higher budget.} Further analysis of importance-guided bit allocation is provided in \Cref{app:importance_allocation}.

\begin{table*}[!tp]
  \centering
  \caption{\rev{Full model architectures of \constant{} and \adaptive{} in encoder and decoder execution order.}}
  \label{tab:appendix_arch}
  \label{tab:adaptive_arch}
  \small
  \setlength{\tabcolsep}{3pt}
  \renewcommand{\arraystretch}{1.04}
  \begin{tabular}{@{}L{0.15\linewidth}L{0.08\linewidth}L{0.245\linewidth}L{0.05\linewidth}L{0.115\linewidth}L{0.29\linewidth}@{}}
    \toprule
    Block & \shortstack[l]{\revheading{Adaptive}\\\revheading{only}} & Main operation & $k/s$ & Channels & Output \\
    \midrule
    \multicolumn{6}{l}{\revheading{Encoder: input views to bitstream}} \\
    \midrule
    \rev{Feature extraction $f$} & \revheading{No} & \rev{MVSplat+ZPressor} & -- & -- & $\feat_v:\;133{\times}H{\times}W$ \\
    \addlinespace[2pt]
    \multirow{3}{=}{\rev{Encoder importance $p_{\mathrm{enc}}$}} & \multirow{3}{*}{\revheading{Yes}} & $\mathrm{C}$ + ReLU & $3/1$ & $134{\to}64$ &  \\
     &  & $\mathrm{C}$ + ReLU & $3/1$ & $64{\to}32$ &  \\
     &  & $\mathrm{C}$ + Sigmoid & $1/1$ & $32{\to}1$ & $S_{\mathrm{enc},v}:\;H{\times}W$ \\
    \addlinespace[2pt]
    \multirow{3}{=}{Analysis $g_a$} & \multirow{3}{*}{\revheading{No}} & $\mathrm{C}$ + GDN & $5/2$ & $133{\to}192$ & $192{\times}H/2{\times}W/2$ \\
     &  & [$\mathrm{C}$ + GDN] ($\times2$) & $5/2$ & $192{\to}192$ & $192{\times}H/8{\times}W/8$ \\
     &  & $\mathrm{C}$ & $5/2$ & $192{\to}320$ & $y_v:\;320{\times}H/16{\times}W/16$ \\
    \addlinespace[2pt]
    \rev{Latent gating} & \revheading{Yes} & \rev{Average-pool $S_{\mathrm{enc},v}$; multiply $y_v$ elementwise} & -- & -- & \rev{Gated $y_v$ (same shape)} \\
    \addlinespace[2pt]
    \multirow{3}{=}{Hyper-analysis $h_a$} & \multirow{3}{*}{\revheading{No}} & $\mathrm{C}$ + LReLU & $3/1$ & $320{\to}192$ & $192{\times}H/16{\times}W/16$ \\
     &  & $\mathrm{C}$ + LReLU & $5/2$ & $192{\to}192$ & $192{\times}H/32{\times}W/32$ \\
     &  & $\mathrm{C}$ & $5/2$ & $192{\to}192$ & $z_v:\;192{\times}H/64{\times}W/64$ \\
    \addlinespace[2pt]
    \rev{Hyper-latent coding} & \revheading{No} & \rev{Entropy bottleneck: encode $z_v$, then locally decode} & -- & $192$ & $B_{z,v},\;\hat z_v$ \\
    \addlinespace[2pt]
    \multirow{3}{=}{Hyper-synthesis $h_s$} & \multirow{3}{*}{\revheading{No}} & $\mathrm{TC}$ + LReLU & $5/2$ & $192{\to}320$ & $320{\times}H/32{\times}W/32$ \\
     &  & $\mathrm{TC}$ + LReLU & $5/2$ & $320{\to}480$ & $480{\times}H/16{\times}W/16$ \\
     &  & $\mathrm{C}$ & $3/1$ & $480{\to}640$ & $\mu_v,\sigma_v:\;320{\times}H/16{\times}W/16$ each \\
    \addlinespace[2pt]
    \rev{Main-latent coding} & \revheading{No} & \rev{Gaussian conditional: encode $y_v$ using $\mu_v,\sigma_v$} & -- & $320$ & $B_{y,v}$ \\
    \addlinespace[2pt]
    \midrule
    \multicolumn{6}{l}{\revheading{Decoder: bitstream and keep ratio to Gaussians}} \\
    \midrule
    \multirow{2}{=}{\rev{Keep-ratio embedding}} & \multirow{2}{*}{\revheading{Yes}} & \rev{128-frequency sinusoidal encoding of $k$} & -- & $256$ &  \\
     &  & \rev{Linear, SiLU, linear} & -- & \shortstack[l]{$256{\to}256$\\${\to}256$} & $e(k):\;256$ \\
    \addlinespace[2pt]
    \rev{Hyper-latent decoding} & \revheading{No} & \rev{Entropy bottleneck: decode $B_{z,v}$} & -- & $192$ & $\hat z_v:\;192{\times}H/64{\times}W/64$ \\
    \addlinespace[2pt]
    \multirow{3}{=}{\rev{Hyper-synthesis $h_s$}} & \multirow{3}{*}{\revheading{No}} & $\mathrm{TC}$ + LReLU & $5/2$ & $192{\to}320$ & $320{\times}H/32{\times}W/32$ \\
     &  & $\mathrm{TC}$ + LReLU & $5/2$ & $320{\to}480$ & $480{\times}H/16{\times}W/16$ \\
     &  & $\mathrm{C}$ & $3/1$ & $480{\to}640$ & $\mu_v,\sigma_v:\;320{\times}H/16{\times}W/16$ each \\
    \addlinespace[2pt]
    \rev{Main-latent decoding} & \revheading{No} & \rev{Gaussian conditional: decode $B_{y,v}$ using $\mu_v,\sigma_v$} & -- & $320$ & $\hat y_v:\;320{\times}H/16{\times}W/16$ \\
    \addlinespace[2pt]
    \multirow{3}{=}{Synthesis $g_s$} & \multirow{3}{*}{\revheading{No}} & $\mathrm{TC}$ + IGDN; \rev{FiLM}$^\dagger$ & $5/2$ & $320{\to}192$ & $192{\times}H/8{\times}W/8$ \\
     &  & [$\mathrm{TC}$ + IGDN; \rev{FiLM}$^\dagger$] ($\times2$) & $5/2$ & $192{\to}192$ & $192{\times}H/2{\times}W/2$ \\
     &  & $\mathrm{TC}$; \rev{FiLM}$^\dagger$ & $5/2$ & $192{\to}133$ & $\qfeat_{k,v}:\;133{\times}H{\times}W$ \\
    \addlinespace[2pt]
    \rev{Gaussian attribute prediction} & \revheading{No} & \rev{MVSplat head; FiLM}$^\dagger$\rev{ on refinement features} & -- & -- & \rev{Raw attributes, depth, density} \\
    \addlinespace[2pt]
    \multirow{3}{=}{\rev{Decoder importance $p_{\mathrm{dec}}$}} & \multirow{3}{*}{\revheading{Yes}} & $\mathrm{C}$ + ReLU & $3/1$ & \rev{$143{\to}64$} &  \\
     &  & $\mathrm{C}$ + ReLU; \rev{FiLM} & $3/1$ & \rev{$64{\to}32$} &  \\
     &  & $\mathrm{C}$ + Sigmoid & $1/1$ & \rev{$32{\to}1$} & $S_{\mathrm{dec},v}(k):\;H{\times}W$ \\
    \addlinespace[2pt]
    \rev{Gaussian adapter} & \revheading{No} & \rev{Convert predicted attributes to world-space Gaussians} & -- & -- & $\hat{\gauss}_{\max}:\;N=VHW$ candidates \\
    \addlinespace[2pt]
    \rev{Gaussian selection} & \revheading{Yes} & \rev{Hard top-$K$ on $S_{\mathrm{dec}}(k)$} & -- & -- & $K=\lfloor kN\rfloor$ \revfix{selected} Gaussians \\
    \addlinespace[2pt]
    \bottomrule
  \end{tabular}
  \par\vspace{3pt}
  {\footnotesize\raggedright
  \rev{Base skips Adaptive-only rows marked Yes. $\dagger$: FiLM is applied only in Adaptive, using $e(k)$. The two $h_s$ entries denote the same learned transform evaluated at each endpoint.}\par
  $\mathrm{C}$/ $\mathrm{TC}$: convolution/transposed convolution; $k/s$: kernel size/stride; GDN/IGDN: generalized divisive normalization/inverse; LReLU: leaky ReLU; FiLM: feature-wise linear modulation.
  \rev{Tensor shapes are per plane $v$, with batch omitted; $\qfeat_{k,v}=\qfeat_v$ for Base. Parallel WLS/iWLS shortcuts are added after each $g_a/g_s$ convolution, before GDN/IGDN or FiLM.}\par
  \rev{The scene bitstream contains $\{B_{z,v}\}$ and $\{B_{y,v}\}$ together with decoder metadata. Keep-ratio conditioning does not enter encoding. Training uses a hard top-$K$ forward mask with a sigmoid straight-through gradient.}\par}
\end{table*}

\begin{table*}[!tp]
  \caption{Training schedule and hyperparameters for the \constant{} and \adaptive{} models with MVSplat+ZPressor.}
  \label{tab:hyperparams}
  \centering
  \small
  \setlength{\tabcolsep}{4pt}
  \renewcommand{\arraystretch}{1.06}
  \begin{tabular}{@{}L{0.20\linewidth}L{0.62\linewidth}L{0.10\linewidth}@{}}
    \toprule
    \multicolumn{3}{c}{(a) Training schedule} \\
    \midrule
    Stage & Configuration & Steps \\
    \midrule
    Stage~1: Generator pretraining
      & Train the MVSplat+ZPressor generator, initializing only its multi-view backbone from pretrained UniMatch~\cite{xu2023unimatch} weights
      & 100k \\
    Stage~2: Distortion warm-up
      & Initialize from Stage~1; train the generator and multi-view feature codec for reconstruction without a rate penalty
      & 50k \\
    Stage~3: Joint R--D training
      & Initialize from Stage~2; jointly train the generator and multi-view feature codec for each $\lambda$
      & 60k/rate \\
    Stage~4: \constant{}
      & Initialize from the corresponding Stage~3 checkpoint, freeze only the multi-view feature-extraction backbone, and optimize all remaining modules
      & 60k/rate \\
    Stage~4: \adaptive{}
      & Initialize independently from the same Stage~3 checkpoint, freeze only the multi-view feature-extraction backbone, and optimize the adaptive model with $e(k)$, latent gating, Gaussian selection, and $\lsim$
      & 60k/rate \\
    \bottomrule
  \end{tabular}

  \vspace{6pt}

  \begin{tabular}{@{}L{0.27\linewidth}L{0.66\linewidth}@{}}
    \toprule
    \multicolumn{2}{c}{(b) Hyperparameters} \\
    \midrule
    Item & Value \\
    \midrule
    Optimizer & Adam~\cite{kingma2015adam}, $\beta=(0.9,0.999)$ \\
    Base learning rate & $2\times10^{-4}$ for Stages 1--2, $1\times10^{-4}$ for Stages 3--4 \\
    Learning-rate schedule & 2k-step linear warm-up in all stages \newline $0.1{\times}$ decay at step 50k in Stages 3--4 \\
    Batch size & Stages 1--2: 4 scenes/GPU \newline Stages 3--4: 1 scene/GPU on 4 GPUs \\
    Training views & 12 input views and 8 target views per scene \\
    Image resolution & $256\times256$ for reported experiments \\
    Distortion loss & MSE + $0.05$ LPIPS \\
    Keep-ratio embedding & Sinusoidal encoding + MLP, 256 dimensions \\
    \revfix{Top-$K$ straight-through gradient} & \revfix{Sigmoid} temperature increases linearly from 1.0 to 30.0 over the first 5k Stage~4 iterations \\
    Importance alignment weight & $\beta=0.1$ for $\lsim$ \\
    Gradient clipping & Global-norm clipping at 0.5 \\
    Hardware & 4 NVIDIA RTX 3090 GPUs (24 GB each) \\
    \bottomrule
  \end{tabular}
  \par\vspace{3pt}
  {\footnotesize Keep-ratio embedding, top-$K$ \revfix{selection}, and importance alignment apply only to \adaptive{}.\par}
\end{table*}

\section{Conclusion}
\label{sec:conclusion}

We presented the \method{} framework as an alternative to generate-then-compress for end-to-end feed-forward 3D scene coding. \constant{} encodes multi-view features that jointly represent geometry and appearance and reconstructs Gaussians from the decoded features at the receiver. This design improves compression efficiency while reducing computation, memory usage, and runtime. Across RE10K, ACID, and DL3DV, it achieves an average BD-rate of $-99.8\%$, roughly halves total GMACs, and provides about $12\times$ end-to-end speedup over the MVSplat+FCGS baseline. Bitrate gains also extend to other Gaussian generators. \adaptive{} enables receiver-side adjustment of the Gaussian count from a single bitstream at each trained rate point, preserving reconstruction quality across the evaluated Gaussian budgets. It maintains higher reconstruction quality than the evaluated post-hoc pruning methods even when retaining only $20\%$ of the Gaussians. These results support the potential of multi-view feature coding for 3D live streaming through joint reductions in bitrate and processing time.

\vfill
\section{Future Work}
\label{sec:future_work}

\revfix{Current feed-forward 3DGS methods are typically evaluated on static scenes with sparse, low-resolution input views and limited camera coverage. Our framework shares these limitations, motivating its extension to higher resolutions and wider camera coverage. Reconstructing and coding dynamic scenes presents a further challenge, particularly in exploiting temporal redundancy for multi-view feature coding.}

\makeatletter
\setlength{\@fptop}{0pt}
\setlength{\@fpsep}{12pt}
\setlength{\@fpbot}{0pt plus 1fil}
\setlength{\@dblfptop}{0pt}
\setlength{\@dblfpsep}{12pt}
\setlength{\@dblfpbot}{0pt plus 1fil}
\setcounter{dbltopnumber}{2}
\makeatother
\newpage
\appendices
\section{Implementation details and additional results}
\label{app:appendix}

\subsection{Full model architecture}
\label{app:architecture}
\label{app:training}

\Cref{tab:appendix_arch} summarizes the complete architectures of \constant{} and \adaptive{}, including the Gaussian generator, multi-view feature codec, and adaptive modules, in encoder and decoder execution order.

\revfix{\Cref{tab:hyperparams} gives the four-stage training schedule and hyperparameters. The frozen-generator variant in \Cref{fig:rd} starts from Stage~2, freezes both the feature extractor and Gaussian prediction head, and trains only the codec for 60k iterations at each rate point. During the first 5k iterations of \adaptive{} Stage~4 training, the lower keep-ratio sampling bound decreases linearly from $1.0$ to $k_{\min}=0.1$. Thereafter, each iteration samples $k\sim\mathcal{U}(k_{\min},1.0)$.}

\begin{table*}[!tp]
  \caption{\reviewrev{Module-level complexity analysis on RE10K at $256\!\times\!256$. GMACs exclude G-PCC.} Parameters count learned weights only, in millions (M); G-PCC has no learned weights. MVSplat+FCGS includes Gaussian generation. Entropy processing includes context networks and learned grid projections. Reconstruction includes $g_s$ and AuxT for \textbf{\constant{}} and attribute synthesis for MVSplat+FCGS.}
  \label{tab:decoder_components}
  \centering
  \small
  \setlength{\tabcolsep}{3pt}
  \begin{tabular}{@{}llrrrr@{}}
    \toprule
    Phase & Component & \multicolumn{2}{c}{GMACs} & \multicolumn{2}{c}{Params. (M)} \\
    \cmidrule(lr){3-4}\cmidrule(l){5-6}
          &           & \shortstack{MVSplat+\\FCGS} & \textbf{\constant{}} & \shortstack{MVSplat+\\FCGS} & \textbf{\constant{}} \\
    \midrule
    \multirow{4}{*}{Encoding}
      & Multi-view feature extraction & 712.42 & \textbf{712.42} & 12.47 & \textbf{12.47} \\
      & Gaussian prediction head & 170.51 & \textbf{0.00} & 0.83 & \textbf{0.00} \\
      & Codec encoding & 678.49 & \textbf{120.29} & 2.54 & \textbf{15.32} \\
      & Total & 1561.41 & \textbf{832.71} & 15.83 & \textbf{27.78} \\
    \midrule
    \multirow{4}{*}{Decoding}
      & Entropy-model processing & 729.19 & \textbf{5.87} & 1.78 & \textbf{8.14} \\
      & Representation reconstruction & 156.72 & \textbf{113.13} & 0.41 & \textbf{4.78} \\
      & Gaussian prediction head & 0.00 & \textbf{170.51} & 0.00 & \textbf{0.83} \\
      & Total & 885.91 & \textbf{289.51} & 2.19 & \textbf{13.75} \\
    \bottomrule
  \end{tabular}
\end{table*}

\begin{table*}[!tp]
  \caption{Raw tensor storage and transmitted bitstream sizes for \constant{}. Panel (a) assumes FP32 storage and six cluster-view planes. \revfix{Panel (b) gives mean complete bitstream sizes over the full RE10K, ACID, and DL3DV test sets.}}
  \label{tab:representation_sizes}
  \centering
  \small
  \setlength{\tabcolsep}{4pt}
  \renewcommand{\arraystretch}{1.12}
  \begin{minipage}{0.8\linewidth}
  \footnotesize
  \begin{tabular*}{\linewidth}{@{}l@{\extracolsep{\fill}}lr@{}}
    \toprule
    \multicolumn{3}{c}{(a) Raw tensor storage} \\
    \midrule
    Tensor & Shape & Size (KiB) \\
    \midrule
    Input images \revfixmath{\{I_i\}}: RE10K/ACID & $8\times3\times256\times256$ & 6144 \\
    Input images \revfixmath{\{I_i\}}: DL3DV & $12\times3\times256\times256$ & 9216 \\
    Multi-view features $F$: all datasets & $6\times133\times256\times256$ & 204\,288 \\
    Main latent $y$: all datasets & $6\times320\times16\times16$ & 1920 \\
    Decoded Gaussians $\hat G$: all datasets & $393\,216\times61$ & 93\,696 \\
    \bottomrule
  \end{tabular*}
  \end{minipage}

  \vspace{5pt}

  \begin{minipage}{0.8\linewidth}
  \footnotesize
  \begin{tabular*}{\linewidth}{@{}l@{\extracolsep{\fill}}rrrr@{}}
    \toprule
    \multicolumn{5}{c}{(b) Mean complete bitstream size (KiB)} \\
    \midrule
    Dataset & $\lambda=0.0256$ & $\lambda=0.1024$ & $\lambda=0.4096$ & $\lambda=1.6384$ \\
    \midrule
    RE10K & 63.2 & 31.7 & 14.2 & 6.5 \\
    ACID & 71.9 & 37.1 & 15.8 & 6.6 \\
    DL3DV & 86.1 & 43.9 & 20.5 & 8.2 \\
    \bottomrule
  \end{tabular*}
  \end{minipage}
  \par\vspace{3pt}
  {\footnotesize The Gaussian buffer contains 3 center coordinates, 9 covariance entries, 48 spherical-harmonic coefficients, and 1 opacity per Gaussian. Only the complete bitstream $B$ is transmitted.}
\end{table*}

\subsection{Evaluation protocols}
\label{app:evaluation_protocols}

For \Cref{fig:rd,tab:bd_complexity}, we retain scenes with finite quality and bitstream-size measurements for every method and operating point, \revfix{including the frozen-generator variant and \adaptive{} at $k=1$.} Rates and quality are arithmetic means over these matched scenes, with identical views across methods. BD-rate is computed from each dataset-average curve using piecewise cubic Hermite interpolation (PCHIP) of log-rate over the common PSNR interval, without extrapolation. For \constant{} versus FCGS, these intervals are 25.552--25.650\,dB on RE10K, 26.439--27.136\,dB on ACID, and 20.622--20.716\,dB on DL3DV. We evaluate all five publicly released FCGS checkpoints, whose reconstructions span a limited PSNR range on these test sets. The cross-dataset summary is the arithmetic mean of the three dataset BD-rates. FCGS uses $\lambda\in\{0.0001,0.0002,0.0004,0.0008,0.0016\}$ in \Cref{fig:rd,fig:backbone_generalization}; TMCV uses attribute QP $\in\{0,8,16,24,32\}$ and position QP~0. For generator transfer, MonoSplat- and TranSplat-based \constant{} models are trained on RE10K with 2 input and 4 target views and evaluated on ACID with 2 input and 3 target views.

\subsection{Complexity, runtime, and representation sizes}
\label{app:runtime_protocol}
\label{app:representation_sizes}

\revfix{Encoding spans input images to bitstream construction, including Gaussian generation for MVSplat+FCGS; decoding ends at Gaussian reconstruction.} Timings include bitstream I/O, G-PCC, and format conversion, but exclude data and model loading.

\revfix{Peak GPU memory includes weights and stage allocations.} GMACs in \Cref{tab:decoder_complexity} count convolution, linear, matrix-product, and attention operations, with one multiply--add counted as one MAC. Rendering, normalization, interpolation, entropy-coder byte operations, G-PCC, and FCGS custom CUDA operations are excluded. \Cref{tab:decoder_components} \reviewrev{reports a separate module-level analysis} using \texttt{fvcore}~\cite{fvcore2022} on RE10K, weighting FCGS branches by their mean routing fractions measured on the sampled RE10K scenes used for this analysis; approximately $98\%$ of Gaussians in this sample are routed to the learned branch.

Subtracting separately measured G-PCC encoding and decoding times \revfix{on the same scenes} reduces MVSplat+FCGS's estimated full-pipeline time from $6.41$\,s to $2.99$\,s per scene, averaged across datasets. All other stages, including geometry preprocessing, format conversion, and rendering, remain included. This is $5.7\times$ the $0.52$\,s required by \constant{}.

\revfix{For reference, \Cref{tab:representation_sizes} reports the raw tensor storage requirements and mean transmitted bitstream sizes of MFGS-Base.}

\begin{figure*}[!tp]
  \centering
  \includegraphics[width=\linewidth]{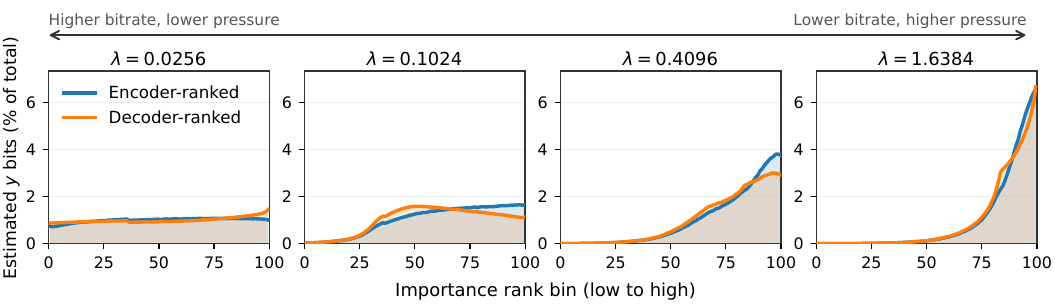}
  \caption{Importance-ranked allocation of estimated \revfix{latent} bits on ACID at $k=0.8$. Each panel uses one rate weight $\lambda$. Curves show the mean percentage of estimated scene-level $y$ bits in each of 100 bins ordered from low to high encoder- or decoder-side importance.}
  \label{fig:importance_bit_allocation}
\end{figure*}

\subsection{Comparison with multi-view image coding}
\label{app:miv}

We evaluate MPEG Immersive Video (MIV)~\cite{boyce2021miv,dziembowski2025mivctc} using TMIV 24.0, VVenC 1.12.0, and VVdeC 2.3.0 with VVC Main10 atlas coding~\cite{bross2021vvc}. Both methods exclude requested target-view camera parameters from the rate and \revfix{use scenes completed at every operating point on RE10K, ACID, and DL3DV.}

\begin{table}[!htbp]
  \caption{\rev{Comparison of multi-view feature coding with MPEG Immersive Video (MIV).}}
  \label{tab:miv_comparison}
  \centering
  \renewcommand{\arraystretch}{1.15}
  \begin{tabular*}{0.85\linewidth}{@{}l@{\extracolsep{\fill}}rr@{}}
    \toprule
    Dataset & \shortstack{\revheading{Common PSNR}\\\revheading{interval (dB)}} & \shortstack{\revheading{MIV BD-rate}\\\revheading{(\%)}} \\
    \midrule
    RE10K & 26.08--27.15 & $+440.92$ \\
    ACID & 26.67--28.92 & $+105.24$ \\
    DL3DV & 20.22--21.55 & $-8.84$ \\
    \bottomrule
  \end{tabular*}
  \par\vspace{4pt}
  {\footnotesize\raggedright
    \rev{BD-rate is computed for MIV relative to \constant{} over the common PSNR interval. Positive values favor \constant{}; negative values favor MIV.}\par}
\end{table}

\constant{} requires lower rates than MIV on RE10K and ACID but a higher rate on DL3DV (\Cref{tab:miv_comparison}). \revfix{On DL3DV, the wider camera baselines limit the reconstruction quality of the feed-forward generator, whereas MIV synthesizes target views directly from the transmitted views.}

\subsection{Importance-guided bit allocation}
\label{app:importance_allocation}

\revfix{We analyze the relationship between predicted importance and estimated coding costs across latent locations (\Cref{fig:importance_bit_allocation}).} The coding cost at each latent location is the sum of the entropy model's negative log-probabilities across channels. We average-pool the encoder and decoder importance maps to the latent resolution, rank locations across all feature planes within each scene, and divide each ranking into 100 equal-sized bins. Each bin's cost is normalized by the scene's total estimated $y$-stream cost before \revfix{averaging over all ACID test scenes} at each rate point and $k=0.8$. \reviewrev{Under both rankings, higher-ranked locations account for a larger share of estimated coding costs, particularly at lower bitrates.}

\subsection{Pruning settings and gating--alignment ablation}
\label{app:pruning_settings_ablation}

\rev{PUP-3DGS uses Fisher-information scores computed with its CUDA rasterizer at the original image resolution and a patch size of 1. LightGaussian uses deterministic pixel-contribution counts weighted by opacity and normalized Gaussian volume raised to $v_{\mathrm{pow}}=0.1$. Both prune the Gaussians reconstructed by \constant{}, without per-scene fine-tuning, distillation, or post-pruning quantization.}

\begin{figure}[!tp]
  \centering
  \includegraphics[width=\linewidth]{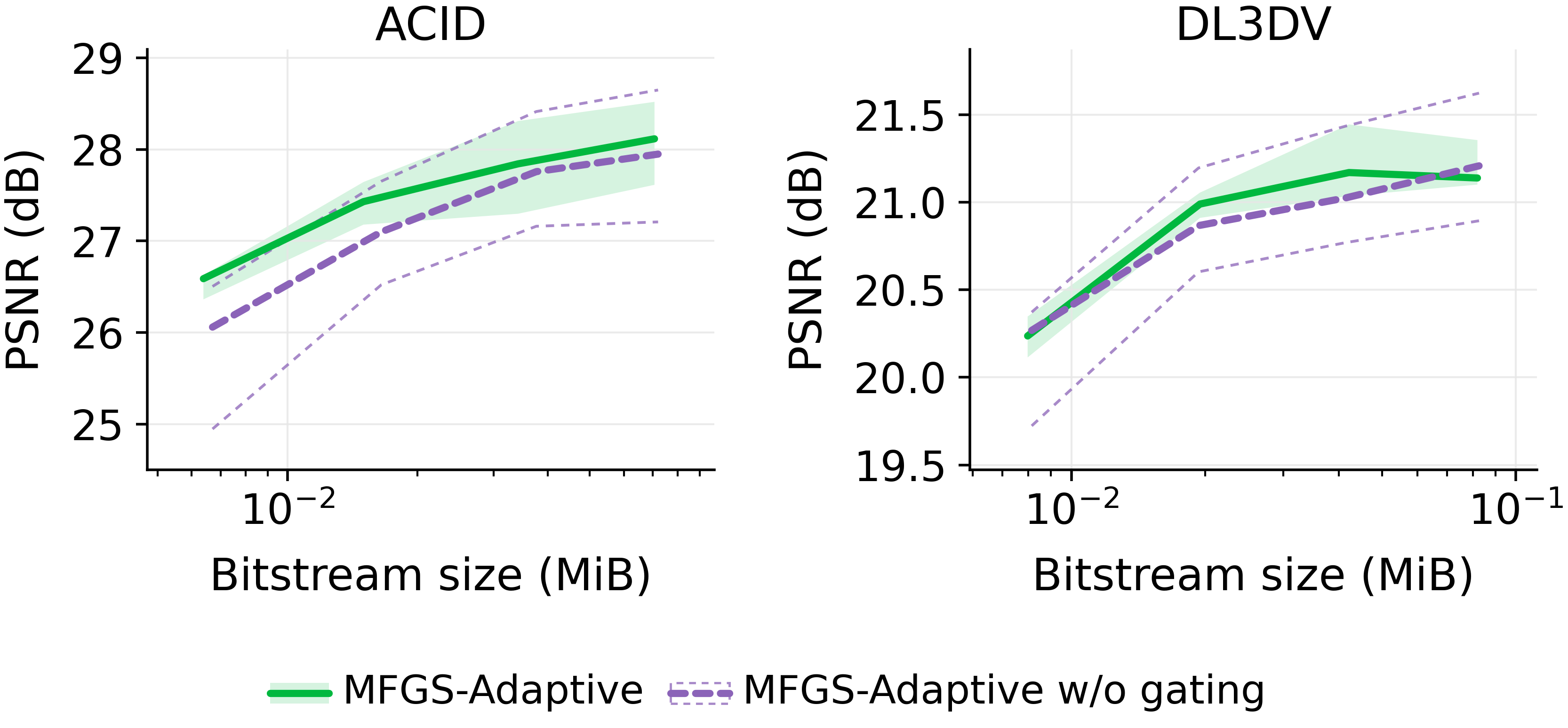}
  \caption{Joint ablation of encoder-side latent gating and importance alignment on ACID and DL3DV. The variant labeled no-gating omits both components. Green shading and thin purple boundary curves indicate the ranges for the full and no-gating models, respectively. These ranges span $k\in\{0.2,0.4,0.6,0.8,1.0\}$; lines connect rate settings at $k=0.4$.}
  \label{fig:gating_ablation}
\end{figure}

The ablation in \Cref{fig:gating_ablation} removes encoder-side gating and the importance-alignment loss while retaining keep-ratio conditioning and decoder-side selection. Both variants use the corresponding Stage~3 checkpoint and the same Stage~4 schedule. Joint gating and alignment narrow the PSNR range across \revfix{$k\in\{0.2,0.4,0.6,0.8,1.0\}$ at each rate point from $0.6$--$1.6$\,dB to $0.1$--$1.0$\,dB on ACID and DL3DV, raise the minimum PSNR across these budgets by up to $1.4$\,dB,} and generally improve the rate--distortion trade-off, with a small reversal at the highest-rate DL3DV point.

\FloatBarrier
\newpage
\raggedbottom
\bibliographystyle{IEEEtran}
\bibliography{references}

@IEEEtranBSTCTL{IEEEexample:BSTcontrol,
  CTLuse_forced_etal = "yes",
  CTLmax_names_forced_etal = "6",
  CTLnames_show_etal = "1"
}

@article{kerbl20233d,
  title={{3D} Gaussian Splatting for Real-Time Radiance Field Rendering},
  author={Kerbl, Bernhard and Kopanas, Georgios and Leimk{\"u}hler, Thomas and Drettakis, George},
  journal={ACM Transactions on Graphics},
  volume={42},
  number={4},
  year={2023}
}

@inproceedings{vaswani2017attention,
  title={Attention Is All You Need},
  author={Vaswani, Ashish and Shazeer, Noam and Parmar, Niki and Uszkoreit, Jakob and Jones, Llion and Gomez, Aidan N. and Kaiser, Lukasz and Polosukhin, Illia},
  booktitle={NeurIPS},
  year={2017}
}

@inproceedings{perez2018film,
  title={{FiLM}: Visual Reasoning with a General Conditioning Layer},
  author={Perez, Ethan and Strub, Florian and de Vries, Harm and Dumoulin, Vincent and Courville, Aaron},
  booktitle={AAAI},
  year={2018}
}

@inproceedings{pixelsplat2023,
  title={{pixelSplat}: {3D} Gaussian Splats from Image Pairs for Scalable Generalizable {3D} Reconstruction},
  author={Charatan, David and Li, Sizhe Lester and Tagliasacchi, Andrea and Sitzmann, Vincent},
  booktitle={CVPR},
  year={2024}
}

@inproceedings{mvsplat2024,
  title={{MVSplat}: Efficient {3D} Gaussian Splatting from Sparse Multi-View Images},
  author={Chen, Yuedong and Xu, Haofei and Zheng, Chuanxia and Zhuang, Bohan and Pollefeys, Marc and Geiger, Andreas and Cham, Tat-Jen and Cai, Jianfei},
  booktitle={ECCV},
  year={2024}
}

@inproceedings{depthsplat2024,
  title={{DepthSplat}: Connecting Gaussian Splatting and Depth},
  author={Xu, Haofei and Peng, Songyou and Wang, Fangjinhua and Blum, Hermann and Barath, Daniel and Geiger, Andreas and Pollefeys, Marc},
  booktitle={CVPR},
  year={2025}
}

@inproceedings{monosplat2025,
  title={{MonoSplat}: Generalizable {3D} Gaussian Splatting from Monocular Depth Foundation Models},
  author={Liu, Yifan and Fan, Keyu and Yu, Weihao and Li, Chenxin and Lu, Hao and Yuan, Yixuan},
  booktitle={CVPR},
  year={2025}
}

@inproceedings{transplat2025,
  title={{TranSplat}: Generalizable {3D} Gaussian Splatting from Sparse Multi-View Images with Transformers},
  author={Zhang, Chuanrui and Zou, Yingshuang and Li, Zhuoling and Yi, Minmin and Wang, Haoqian},
  booktitle={AAAI},
  year={2025}
}

@inproceedings{zpressor2025,
  title={{ZPressor}: Bottleneck-Aware Compression for Scalable Feed-Forward {3DGS}},
  author={Wang, Weijie and Chen, Donny Y. and Zhang, Zeyu and Shi, Duochao and Liu, Akide and Zhuang, Bohan},
  booktitle={NeurIPS},
  year={2025}
}

@article{re10k2018,
  title={Stereo Magnification: Learning View Synthesis Using Multiplane Images},
  author={Zhou, Tinghui and Tucker, Richard and Flynn, John and Fyffe, Graham and Snavely, Noah},
  journal={ACM Transactions on Graphics},
  volume={37},
  number={4},
  pages={65:1--65:12},
  year={2018},
  doi={10.1145/3197517.3201323}
}

@inproceedings{acid2020,
  title={Infinite Nature: Perpetual View Generation of Natural Scenes from a Single Image},
  author={Liu, Andrew and Tucker, Richard and Jampani, Varun and Makadia, Ameesh and Snavely, Noah and Kanazawa, Angjoo},
  booktitle={ICCV},
  year={2021}
}

@inproceedings{dl3dv2024,
  title={{DL3DV-10K}: A Large-Scale Scene Dataset for Deep Learning-Based {3D} Vision},
  author={Ling, Lu and Sheng, Yichen and Tu, Zhi and Zhao, Wentian and Xin, Cheng and Wan, Kun and Yu, Lantao and Guo, Qianyu and Yu, Zixun and Lu, Yawen and others},
  booktitle={CVPR},
  year={2024}
}

@inproceedings{balle2018variational,
  title={Variational Image Compression with a Scale Hyperprior},
  author={Ball{\'e}, Johannes and Minnen, David and Singh, Saurabh and Hwang, Sung Jin and Johnston, Nick},
  booktitle={ICLR},
  year={2018}
}

@inproceedings{minnen2018joint,
  title={Joint Autoregressive and Hierarchical Priors for Learned Image Compression},
  author={Minnen, David and Ball{\'e}, Johannes and Toderici, George},
  booktitle={NeurIPS},
  year={2018}
}

@inproceedings{cheng2020learned,
  title={Learned Image Compression with Discretized Gaussian Mixture Likelihoods and Attention Modules},
  author={Cheng, Zhengxue and Sun, Heming and Takeuchi, Masaru and Katto, Jiro},
  booktitle={CVPR},
  year={2020}
}

@inproceedings{li2025disentangled,
  title={On Disentangled Training for Nonlinear Transform in Learned Image Compression},
  author={Li, Han and Li, Shaohui and Dai, Wenrui and Cao, Maida and Kan, Nuowen and Li, Chenglin and Zou, Junni and Xiong, Hongkai},
  booktitle={ICLR},
  year={2025}
}

@inproceedings{hac2024,
  title={{HAC}: Hash-Grid Assisted Context for {3D} Gaussian Splatting Compression},
  author={Chen, Yihang and Wu, Qianyi and Lin, Weiyao and Harandi, Mehrtash and Cai, Jianfei},
  booktitle={ECCV},
  year={2024}
}

@article{hacpp2025,
  title={{HAC++}: Towards 100X Compression of {3D} Gaussian Splatting},
  author={Chen, Yihang and Wu, Qianyi and Lin, Weiyao and Harandi, Mehrtash and Cai, Jianfei},
  journal={IEEE Transactions on Pattern Analysis and Machine Intelligence},
  volume={47},
  number={11},
  pages={10210--10226},
  year={2025},
  doi={10.1109/TPAMI.2025.3594066}
}

@inproceedings{contextgs2024,
  title={{ContextGS}: Compact {3D} Gaussian Splatting with Anchor Level Context Model},
  author={Wang, Yufei and Li, Zhihao and Guo, Lanqing and Yang, Wenhan and Kot, Alex C. and Wen, Bihan},
  booktitle={NeurIPS},
  year={2024}
}

@inproceedings{rdogaussian2024,
  title={End-to-End Rate-Distortion Optimized {3D} Gaussian Representation},
  author={Wang, Henan and Zhu, Hanxin and He, Tianyu and Feng, Runsen and Deng, Jiajun and Bian, Jiang and Chen, Zhibo},
  booktitle={ECCV},
  year={2024}
}

@inproceedings{compgs2024,
  title={{CompGS}: Smaller and Faster Gaussian Splatting with Vector Quantization},
  author={Navaneet, K. L. and Meibodi, Kossar Pourahmadi and Koohpayegani, Soroush Abbasi and Pirsiavash, Hamed},
  booktitle={ECCV},
  year={2024}
}

@inproceedings{lightgaussian2023,
  title={{LightGaussian}: Unbounded {3D} Gaussian Compression with 15x Reduction and 200+ FPS},
  author={Fan, Zhiwen and Wang, Kevin and Wen, Kairun and Zhu, Zehao and Xu, Dejia and Wang, Zhangyang},
  booktitle={NeurIPS},
  year={2024}
}

@inproceedings{fcgs2025,
  title={Fast Feedforward {3D} Gaussian Splatting Compression},
  author={Chen, Yihang and Wu, Qianyi and Li, Mengyao and Lin, Weiyao and Harandi, Mehrtash and Cai, Jianfei},
  booktitle={ICLR},
  year={2025}
}

@article{schwarz2019emerging,
  title={Emerging {MPEG} Standards for Point Cloud Compression},
  author={Schwarz, Sebastian and Preda, Marius and Baroncini, Vittorio and Budagavi, Madhukar and C{\'e}sar, Pablo and Chou, Philip A. and Cohen, Robert A. and Krivoku{\'c}a, Maja and Lasserre, S{\'e}bastien and Li, Zhu and Llach, Joan and Mammou, Khaled and Mekuria, Rufael and Nakagami, Ohji and Siahaan, Ernestasia and Tabatabai, Ali J. and Tourapis, Alexis M. and Zakharchenko, Vladyslav},
  journal={IEEE Journal on Emerging and Selected Topics in Circuits and Systems},
  volume={9},
  number={1},
  pages={133--148},
  year={2019},
  doi={10.1109/JETCAS.2018.2885981}
}

@misc{mpeggsc2025,
  title  = {{[GSC]} Reference Software Proposal for {3D} {G}aussian Splat Compression Using Video in {V3C} Bitstream},
  author = {Ricard, Julien and Teniou, Gilles and Wenger, Stephan},
  howpublished = {{ISO/IEC} {JTC} 1/{SC} 29/{WG} 7, Doc.\ m73254, 151st {MPEG} Meeting, Daejeon, Republic of Korea},
  month  = jun,
  year   = {2025}
}

@inproceedings{pup3dgs2024,
  title={{PUP 3D-GS}: Principled Uncertainty Pruning for {3D} Gaussian Splatting},
  author={Hanson, Alex and Tu, Allen and Singla, Vasu and Jayawardhana, Mayuka and Zwicker, Matthias and Goldstein, Tom},
  booktitle={CVPR},
  year={2025}
}

@inproceedings{ecosplat2025,
  title={{EcoSplat}: Efficiency-controllable Feed-forward {3D} Gaussian Splatting from Multi-view Images},
  author={Park, Jongmin and Bui, Minh-Quan Viet and {Gonzalez Bello}, Juan Luis and Moon, Jaeho and Oh, Jihyong and Kim, Munchurl},
  booktitle={CVPR},
  year={2026}
}

@inproceedings{zhang2018lpips,
  title={The Unreasonable Effectiveness of Deep Features as a Perceptual Metric},
  author={Zhang, Richard and Isola, Phillip and Efros, Alexei A. and Shechtman, Eli and Wang, Oliver},
  booktitle={CVPR},
  year={2018}
}

@article{xu2023unimatch,
  title={Unifying Flow, Stereo and Depth Estimation},
  author={Xu, Haofei and Zhang, Jing and Cai, Jianfei and Rezatofighi, Hamid and Yu, Fisher and Tao, Dacheng and Geiger, Andreas},
  journal={IEEE Transactions on Pattern Analysis and Machine Intelligence},
  volume={45},
  number={11},
  pages={13941--13958},
  year={2023},
  doi={10.1109/TPAMI.2023.3298645}
}

@techreport{bjontegaard2001calculation,
  title={Calculation of Average {PSNR} Differences between {RD}-Curves},
  author={Bj{\o}ntegaard, Gisle},
  institution={{ITU-T} Video Coding Experts Group ({VCEG})},
  type={Meeting contribution},
  number={VCEG-M33},
  address={Austin, Texas, USA},
  year={2001}
}

@article{boyce2021miv,
  title={{MPEG} Immersive Video Coding Standard},
  author={Boyce, Jill M. and Dor{\'e}, Renaud and Dziembowski, Adrian and Fleureau, Julien and Jung, Joel and Kroon, Bart and Salahieh, Basel and Malamal Vadakital, Vinod Kumar and Yu, Lu},
  journal={Proceedings of the IEEE},
  volume={109},
  number={9},
  pages={1521--1536},
  year={2021},
  doi={10.1109/JPROC.2021.3062590}
}

@article{bross2021vvc,
  title={Overview of the Versatile Video Coding ({VVC}) Standard and Its Applications},
  author={Bross, Benjamin and Wang, Ye-Kui and Ye, Yan and Liu, Shan and Chen, Jianle and Sullivan, Gary J. and Ohm, Jens-Rainer},
  journal={IEEE Transactions on Circuits and Systems for Video Technology},
  volume={31},
  number={10},
  pages={3736--3764},
  year={2021},
  doi={10.1109/TCSVT.2021.3101953}
}

@techreport{dziembowski2025mivctc,
  title={Common Test Conditions for {MPEG} Immersive Video},
  author={Dziembowski, Adrian and Kroon, Bart},
  institution={{ISO/IEC JTC 1/SC 29/WG 4}},
  number={N0659},
  year={2025},
  url={https://www.multimedia.edu.pl/publications/N0659-Common-test-conditions-for-MPEG-immersive-video.pdf}
}

@misc{fvcore2022,
  author={{FAIR}},
  title={{fvcore}: A Lightweight Core Library for Computer Vision},
  howpublished={Software package, version 0.1.5.post20221221},
  year={2022},
  url={https://pypi.org/project/fvcore/0.1.5.post20221221/}
}

@inproceedings{kingma2015adam,
  title={Adam: A Method for Stochastic Optimization},
  author={Kingma, Diederik P. and Ba, Jimmy},
  booktitle={ICLR},
  year={2015}
}

\end{document}